\documentclass{article}

\usepackage{PRIMEarxiv}
\usepackage[utf8]{inputenc} 
\usepackage[T1]{fontenc}    
\usepackage{hyperref}       
\usepackage{url}            
\usepackage{booktabs}       
\usepackage{amsfonts}       
\usepackage{nicefrac}       
\usepackage{microtype}      
\usepackage{lipsum}
\usepackage{fancyhdr}       
\usepackage{graphicx}       
\graphicspath{{media/}}     
\usepackage{amsmath}
\usepackage{amssymb}
\usepackage{xcolor}
\usepackage{float}
\usepackage{subcaption}
\usepackage{todonotes}
\usepackage{tikz,pgfplots,pgf}
\usetikzlibrary{matrix,shapes,arrows,positioning}
\usepackage{units}
\usepackage{upgreek}
\usepackage{comment}
\usepackage[titletoc, title]{appendix}
\usepackage{multirow}
\usepackage{tikz}
\usepackage{caption}

\title{Enhancing material behavior discovery using embedding-oriented Physically-Guided Neural Networks with Internal Variables.
}

\author{
Rubén Muñoz-Sierra \\
  Arag\'on Institute of Engineering Research (I3A);\\
  University of Zaragoza;\\
  Mariano Esquillor, S/N,  50018 Zaragoza, Spain. \\
  Institute for Health Research Aragon (IISA);\\
  Avenida San Juan Bosco, 13,  50009 Zaragoza, Spain.\\
  \texttt{rmunnoz@iisaragon.es} \\
\And
  Manuel Doblaré \\
  Mechanical Engineering Department;\\
  Arag\'on Institute of Engineering Research (I3A);\\
  University of Zaragoza;\\
  Mariano Esquillor, S/N,  50018 Zaragoza, Spain;\\
  Nanjing Tech University; \\
  Jiangsu Province, China.\\
  \texttt{mdoblare@unizar.es} \\
\And
  Jacobo Ayensa-Jim\'enez\thanks{Corresponding author} \\
  Institute for Health Research Aragon (IISA);\\
  Avenida San Juan Bosco, 13,  50009 Zaragoza, Spain.\\
  Department of Mechanical Engineering, Universidad Politécnica de Madrid (UPM);\\
  José Gutiérrez Abascal, 2,  28006 Madrid, Spain.\\
  \texttt{jacobo.ayensa@upm.es} \\
}

\begin{document}
\newcommand\norm[1]{\left\lVert#1\right\rVert}
\newcommand{\bs}[1]{\boldsymbol{#1}}
\newcommand{\ssf}[1]{\mathsf{#1}}
\tikzset{%
  every neuron/.style={
    circle,
    draw,
    minimum size=0.5cm
  },
  neuron missing/.style={
    draw=none, 
    scale=3,
    text height=0.2cm,
    execute at begin node=\color{black}$\vdots$
  },
}

\maketitle

\begin{abstract}

Physically Guided Neural Networks with Internal Variables are Scientific Machine Learning tools that use only observable data for training and and have the capacity to unravel internal state relations. They incorporate physical knowledge both by prescribing the model architecture and using loss regularization, thus endowing certain specific neurons with a physical meaning as internal state variables. Despite their potential, these models face challenges in scalability when applied to high-dimensional data such as fine-grid spatial fields or time-evolving systems.

In this work, we propose some enhancements to the PGNNIV framework that address these scalability limitations through reduced-order modeling techniques. Specifically, we introduce alternatives to the original decoder structure using spectral decomposition (e.g., Fourier basis), Proper Orthogonal Decomposition (POD), and pretrained autoencoder-based mappings. These surrogate decoders offer different trade-offs between computational efficiency, accuracy, noise tolerance, and generalization, while improving drastically the scalability. Additionally, we integrate model reuse via transfer learning and fine-tuning strategies to exploit previously acquired knowledge, supporting efficient adaptation to novel materials or configurations. This also significantly reduces training time while maintaining or improving model performance. To illustrate these techniques, we use a representative case governed by the nonlinear diffusion equation, using only observable data.

The results demonstrate that the enhanced PGNNIV framework successfully identifies the underlying constitutive state equations while maintaining high predictive accuracy. It also improves robustness to noise, mitigates overfitting, and reduces computational time. The proposed techniques can be tailored to various scenarios depending on data availability, computational resources, and specific modeling objectives, overcoming scalability challenges in all scenarios.

\end{abstract}


\section{Introduction} \label{secc::introduction}



In recent years, the field of Simulation-Based Engineering and Sciences (SBES) \cite{chinesta2020virtual,chinesta2022empowering}, and especially Computational Mechanics \cite{moya2024computational,herrmann2024deep}, has been profoundly impacted by the emergence of Scientific Machine Learning (SciML), also known as Physics-Informed Machine Learning (PIML) \cite{karniadakis2021physics}. This approach seeks to create a powerful synergy between the cutting-edge, scalable methods of Machine Learning (ML) and the foundational laws of physics that have been the cornerstone of scientific discovery for centuries. The core principle of SciML is the integration of physical knowledge into AI models \cite{muther2023physical,jiao2024ai}, moving beyond "black-box" approaches like Deep Learning (DL) by enriching them with physical information \cite{kim2021knowledge}. This novel paradigm masterfully combines inductive, data-driven reasoning with deductive, physics-based logic, catalyzing a convergence between the third and fourth pillars of science: computational sciences \cite{quarteroni2009mathematical} and machine learning \cite{von2020introducing}, respectively. The widespread adoption of PIML has led to significant advancements across a multitude of scientific and engineering fields. Its applications are diverse, ranging from improving reliability and structural integrity assessments \cite{zhu2023physics,xu2023physics} to enhancing anomaly detection and system monitoring \cite{cross2022physics,wu2024physics}. Furthermore, PIML has proven invaluable in refining weather and climate models \cite{kashinath2021physics,bracco2025machine} and in addressing complex challenges within the realms of solid and fluid mechanics \cite{jin2023recent,sharma2023review,faroughi2024physics}.

A variety of strategies have been developed to incorporate prior physical knowledge into Neural Networks (NNs) for solving problems governed by Partial Differential Equations (PDEs). These methods enable the extraction of physically consistent insights from observational data and are applied to a wide array of tasks, including the computation of solutions of equations, operator learning, or the discovery of underlying properties, among others. One widely adopted technique is Physics-Informed Neural Networks (PINNs) \cite{cuomo2022scientific, raissi2019physics}, which embeds governing equations directly into the loss function. By penalizing the model for deviations from initial and boundary conditions, as well as the PDE residuals at selected points in the domain, PINNs effectively constrain the solution space to physically plausible outcomes. This method has demonstrated to be effective in solving problems obtaining the data-driven solution of PDEs, and in the data-driven discovery of specific properties or parameters of the physical system, when measurement data is also incorporated into the loop. 
In constrast, Structure-Preserving Neural Networks (SPNNs) \cite{hernandez2021structure, hernandez2022thermodynamics} are architecturally designed to inherently respect fundamental principles inherent in physical systems, such as the conservation of energy, symmetry, geometric properties, or the principles of thermodynamics, ensuring the preservation of these properties over time. Using observed data, SPNNs can predict the future state of the system without requiring explicit knowledge of its governing equations. 
Another increasingly popular approach is to learn operators that map between infinite-dimensional function spaces, rather than relying uniquely on finite-dimensional approximations. Neural Operators \cite{kovachki2023neural} achieve this by learning global mappings between discretized functions, often using architectures like the Fourier Neural Operator (FNO) \cite{li2020fourier}, which applies spectral convolutions to efficiently capture long-range dependencies and generalize across resolutions. On the other hand, Deep Operator Networks (DeepONets)\cite{lu2021learning}, represent the operator by means of a two-network structure: a \emph{branch network} that encodes the input functions at a set of sensor points, and a \emph{trunk network} that processes the coordinates for the output locations. This architecture allows the model to generalize across both input functions and evaluation domains. Finally, NN-EUCLID \cite{thakolkaran2022nn} is a methodology that combines DL with the EUCLID (Efficient Unsupervised Constitutive Law Identification and Discovery) framework \cite{flaschel2022discovering, flaschel2023automated}. EUCLID is an unsupervised, physics-informed framework for data-driven discovery of constitutive laws, informed by full-field displacement and reaction force measurements. It employs sparse regression to identify an appropriate material model from a potentially large space of candidate formulations. In its Neural Network-based variant, NN-EUCLID, the constitutive model is represented by input-convex neural networks (ICNNs), which are architecturally constrained to learn convex functions, thereby inherently enforcing physical and thermodynamic consistency.

In addition to the aforementioned approaches, a hybrid method known as Physically-Guided Neural Networks with Internal Variables (PGNNIV) \cite{ayensa2021prediction,ayensa2025predicting,munoz2025application} combines a specific network structure with loss regularization based on physical constraints. In this framework, the Physical knowledge is introduced both using penalty terms at the loss regularization and at the network architecture level, by explicitly distinguishing between measurable and non-measurable (internal) variables. Universal physical laws act as constraints, allowing the values of certain neurons to be interpreted as the internal state variables of the system. When the architecture is properly designed, this endows the network with unraveling capacity \cite{ayensa2022understanding,ayensa2025predicting}. Indeed, only observable data are used to train the network, while the internal state equations may be extracted as a result of the training process. This obviates the need to preassume the specific form of the internal state model while still ensuring that the solutions are physically consistent \cite{ayensa2025predicting,munoz2025application}. Beyond this explanatory character, PGNNIVs improve conventional ANNs as they exhibit faster convergence, lower data requirements, and present additional noise filtering capacity \cite{ayensa2021prediction}. 

Despite their strengths, PGNNIVs present many limitations that have been extensively discussed \cite{ayensa2021prediction}. A primary challenge stems from the predictive component of the network often operating at a granular level (e.g., pixel or voxel), which can lead to an explosion in the number of output variables, particularly for high-resolution images or unstructured computational meshes \cite{ayensa2025predicting,munoz2025application}. Moreover, when extending the method to higher dimensions (for instance by increasing the number of channels or by applying it to time evolution problems), the method is doomed to the curse of dimensionality \cite{gonzalez2010recent,poggio2017and}. Another significant drawback of PGNNIVs is that they are data-hungry, both in data quantity (high sample size) \cite{ayensa2021prediction,ayensa2025predicting} and in data quality (high data variability) \cite{ayensa2025predicting,munoz2025application}. 

Finally, it is important to note that the main purpose of a PGNNIV is to unravel constitutive behaviors. This raises a fundamental question,  whether it is possible to leverage the already knowledge about a given physical system (understood as an experimental platform and a particular specimen) to apply it to a different material behavior, without the need of facing the problem from scratch.

In this paper, we proposed different strategies, all based on embedding representations of the predictive component, that partially solve each of these challenges. Inspired by recent advancements \cite{anowar2021conceptual,jia2022feature}, the central idea is that an embedding representation of the predictive component is able to encode in a latent space the information about the experimental or computational setup. This approach yields a more compact and efficient representation of the system, which can significantly reduce data requirements and mitigate the curse of dimensionality, all while preserving the constitutive behavior discovery capabilities of the explanatory component. Furthermore, the modular nature of the PGNNIV allows for the different components, which encode knowledge at various levels, to be exported and reused, thereby harnessing the benefits of Transfer Learning (TL) to conserve computational resources.

The primary objective of this work is to enhance the PGNNIV methodology for continuum problems by improving its computational efficiency, addressing its scalability limitations, and increasing its robustness to noisy data. This is achieved through the integration of low-order representations, which reduce model complexity and, consequently, training demands, while maintaining or even enhancing the model's predictive accuracy and its ability to uncover underlying physical properties across diverse datasets and noise levels. 

The structure of the paper is as follows: firstly, in section \ref{secc::theo_background} we introduce the problem to solve and the PGNNIV methodology, and we outline the main ideas of the work. In this section we also present the different techniques proposed, as they are the dimensionality reduction techniques or knowledge transfer, and we evaluate the model complexity, as well as how the acquired knowledge can be leveraged to discover constitutive behaviors for new materials. In section \ref{secc::methods}, we describe the general framework for studying continuum physics within the PGNNIV approach (reformulating the mathematical foundations of continuum mechanics using the formalism of ANNs), we particularize the different embedding technique to our problem in hands and we describe the computational details associated with data generation and the model architecture. In section \ref{secc::results} we present some numerical experiments that support our claims, illustrating the performance of the different approaches proposed, under different scenarios. These results are discussed in section \ref{secc::discussion}, where we analyze the effect of the dimension of the latent space, the impact of the sample size and data uncertainty. Finally, in section \ref{secc::discussion} we provide a discussion and we summarize the main conclusions of the study. To ensure the reproducibility of our research—a fundamental tenet of Scientific Machine Learning—we provide an open-source GitHub repository containing the complete source code and all computational experiments for review and further exploration: \url{https://github.com/rmunozTMELab/Embedding-Oriented-PGNNIV}

\section{Theoretical background} \label{secc::theo_background}

\subsection{Physically-Guided Neural Networks with Internal Variables} \label{subsecc::pgnniv}

The use of PGNNIVs is a recently introduced methodology in the context of SciML. In this approach, universal physical laws are used as constraints in a given neural network, in such a way that some neuron values can be interpreted as internal state variables of the system. 

Let us consider a certain physical problem defined by a set of (possibly nonlinear) partial differential equations (PDEs) which can usually be split into two main groups, universal physical laws and constitutive equations, and completed with appropriate initial and/or boundary conditions:

\begin{subequations}\label{eq::eq_fundamental}
\begin{align}
    \mathcal{F}[\bs{u},\bs{v}] &= \bs{f}, \label{eq::eq_fundamental1}\\
    \mathcal{H}[\bs{u},\bs{v}] &= \bs{0}. \label{eq::eq_fundamental2}\\
    \mathcal{G}[\bs{u},\bs{v},\bs{g}] &= \bs{0}, \label{eq::eq_fundamental3}
\end{align}
\end{subequations}

$\mathcal{F}$, $\mathcal{H}$ and $\mathcal{G}$ are differential operators that relate the different tensor field quantities, $\bs{u}$, $\bs{v}$, $\bs{g}$ and $\bs{f}$, also with their own tensor character. The functionals $\mathcal{F}$ and $\mathcal{G}$ respectively represent the whole set of universal laws associated with the problem in hands, as well as particular geometric and environmental constraints respectively, whereas $\mathcal{H}$ represents all (possibly unknown) internal state or constitutive relations of the problem. Splitting of the problem in these two sets of PDEs systems also drives to the distinction between the two kind of unknown fields: the essential measurable fields, $\bs{u}$, and the internal state fields, $\bs{v}$, that are particular to each continuum based field theory. In many contexts, the underlying theory is formulated such that Eq. (\ref{eq::eq_fundamental2}) may be expressed in the form $\bs{v} = \mathcal{H}(\bs{u})$. For instance, in solid mechanics, $\mathcal{F}$ encodes mass, momentum and energy conservation equations, while $\mathcal{H}$ expresses the material-dependent constitutive relations between stresses (the internal variables) and strains (that can be obtained from the deformation mapping) \cite{ayensa2025predicting}. Similarly, for heat transfer problems, $\mathcal{F}$ is related to energy conservation, whereas $\bs{v} = \mathcal{H}(\bs{u})$ is the constitutive law for thermal conduction (where $\bs{u} = u$ is the temperature scalar field, and $\bs{v} = \bs{q}$ is the energy flux vector), for instance the Fourier's law \cite{munoz2025application}.

Setting aside the physical problem, we can think of a ML problem consisting of a collection of input-output data $\mathcal{D} = \{(\mathbf{x}^i,\mathbf{y}^i), \ i=1,\ldots,D\}$ whose aim is to learn the underlying implicit relationship $\mathbf{y} = \bs{Y}(\mathbf{x})$ (note that we use the non-italic notation to represent vector variables without necessarily having a physical interpretation). To face this problem it is possible to set-up a DL regression model relating $\mathbf{x}$ and $\mathbf{y}$ that may be expressed as $\hat{\mathbf{y}} = \mathsf{Y}(\mathbf{x})$ (note that the use of sans serif notation indicates that there is a DL model relating these two variables). Once the error $\mathbf{e} = \hat{\mathbf{y}} - \mathbf{y}$ is defined, the construction of the model $\mathsf{Y}$ is performed by solving a minimization problem, as in conventional ML approaches. We usually define a loss function related to the norm of the error for the whole learning dataset $\mathcal{D}$, for instance $\mathcal{L}_\mathrm{data} = \sum_{i=1}^D\|\mathbf{e}^i\|^2$, with $\mathbf{e}^i = \mathsf{Y}(\mathbf{x}^i) - \mathbf{y}^i$.

Thus, it is possible to couple the physics and the ML problem. Selecting appropriately the input-output pairs, $\mathbf{x}$ and $\mathbf{y}$ respectively, and choosing the correct physical constraints, we can define a problem in terms of the PGNNIV as
\begin{subequations}\label{eq:PGNNIV_problem}
    \begin{align}
        \mathbf{x} &= \mathcal{I}(\bs{u}, \bs{f}, \bs{g}) \label{eq:PGNNIV_problem_c}, \\
        \mathbf{y} &= \mathcal{O}(\bs{u}, \bs{f}, \bs{g}) \label{eq:PGNNIV_problem_d}, \\
        \mathbf{y} &= \bs{Y}(\mathbf{x}) \label{eq:PGNNIV_problem_a}, \\
        \bs{v} &= \mathcal{H}(\bs{u}) \label{eq:PGNNIV_problem_b}, \\
        \mathcal{C}(\bs{u}, \bs{v}, \bs{f}, \bs{g}) &= 0 \label{eq:PGNNIV_problem_e}.
    \end{align}
\end{subequations}
where $\bs{u}$, $\bs{f}$ , and $\bs{g}$ are the measurable fields of the problem. Besides, input $\mathbf{x} \in \Omega$ and output $\mathbf{y}$ will be defined depending on which relation $\mathbf{x} \mapsto \mathbf{y}$ wants to be predicted, and are defined by means of the functionals $\mathcal{I}$ and $\mathcal{O}$ respectively. In addition, $\mathcal{C}$ are the physical constraints, related to the relations given by $\mathcal{F}$ and $\mathcal{G}$. Finally, there are two Deep Neural Networks (DNN) components: 
\begin{itemize}
    \item $\bs{Y}$ is the predictive model, whose aim is to infer accurate values for the output variables in a certain domain of the input $\Omega$, that is, to surrogate the relation $\mathbf{x} \mapsto \mathbf{y}$ on $\Omega$.
    \item $\mathcal{H}$ is the explanatory model, whose objective is to unravel the hidden physics of the relation $\bs{u} \mapsto \bs{v}$.
\end{itemize} 
Note that all the involved elements are spatial fields so the Eqs. (\ref{eq:PGNNIV_problem}) have to be spatially discretized obtaining
\begin{subequations}\label{eq:PGNNIV_problem_discretized}
    \begin{align}
        \mathbf{x} &= \bs{I}(\bs{u}^{(N)}, \bs{f}^{(N)}, \bs{g}^{(N)}), \\
        \mathbf{y} &= \bs{O}(\bs{u}^{(N)}, \bs{f}^{(N)}, \bs{g}^{(N)}), \\
        \mathbf{y} &= \bs{Y}(\mathbf{x}), \\
        \bs{v} &= \bs{H}(\bs{u}^{(N)}), \\
        \bs{C}(\bs{u}^{(N)}, \bs{v}^{(N)}, \bs{f}^{(N)}, \bs{g}^{(N)}) &= 0,
    \end{align}
\end{subequations}
where now $\bs{u}^{(N)}$, $\bs{v}^{(N)}$, $\bs{f}^{(N)}$ and $\bs{g}^{(N)}$ are vectors associated with their respective fields and $\bs{I}$, $\bs{O}$ and $\bs{C}$ are vector functions. Now, both functions $\bs{Y}$ and $\bs{H}$ may be replaced by DNN surrogates, $\mathsf{Y}$ and $\mathsf{H}$. For instance, $\bs{u}^{(N)} = (\bs{u}(\bs{r}_j))_{j=1}^N$ for some $\bs{r}_j \in \mathcal{R}$, $j=1,\ldots,N$.

\begin{figure}[h!]
    \centering
    \includegraphics[width=1\linewidth]{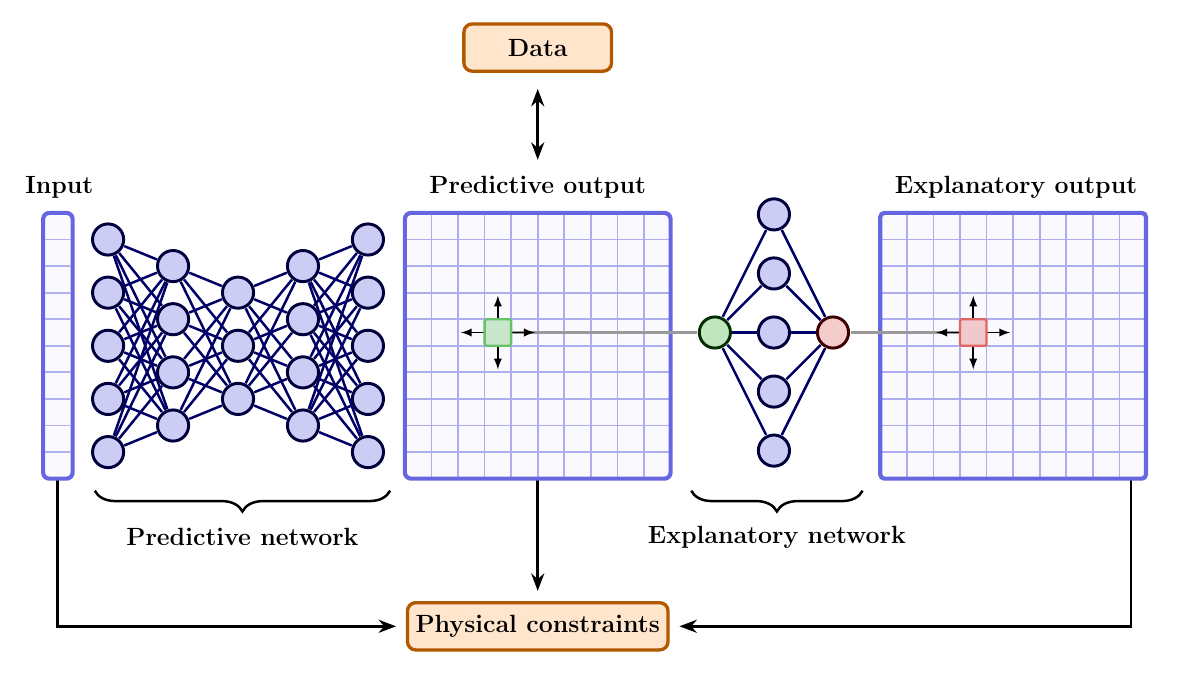}
    \caption{\textbf{Physically-Guided Neural Network baseline model architecture.} The predictive network has a bottleneck-like structure while the explanatory network has a sparse inverted bottleneck-like structure. In addition, physical constranints are imposed as regularization terms in the loss function.}
    \label{fig:methods_pgnniv_baseline_architecture}
\end{figure}

The PGNNIV is a particular ML architecture built by combining two coupled NNs: a predictive network $\mathsf{Y}$ and an explanatory network $\mathsf{H}$. These two networks are trained simultaneously, and their learning processes depend on each other. The predictive network $\mathsf{Y}$ is trained in a supervised manner using known data, the pairs $(\mathbf{x},\mathbf{y})$, where it learns to produce accurate outputs by mapping known input-output pairs. For instance, it receives as input the boundary conditions of a physical system and its aim is to learn and reproduce a field throughout the domain, such as displacement fields in solid mechanics or temperature fields in heat transfer problems. In other words, it learns the solution of the PDE that governs the system. Meanwhile, the explanatory network $\mathsf{H}$ is trained in a self-supervised manner without any extra data, only by constraining its solution space to the one that is physics-consistent via the constraint $\bs{C}$. It is fed with the solutions predicted by the predictive network and seeks to uncover hidden variables, discovering the constitutive laws of the material, and capturing possibly nonlinear behaviors. By integrating both networks, the model not only achieves accurate field predictions, but also provides explainabilty through the discovery of underlying physical properties hidden in the data. A schematic representation of the baseline PGNNIV model is shown in Fig. \ref{fig:methods_pgnniv_baseline_architecture}.

\subsection{Latent space embedding}\label{subsec::latent_space_embedding}


Let us first focus on the predictive component $\mathsf{Y}$ whose aim is to learn the real function $\bs{Y}$.  The value of the output variable is obtained by some image acquisition techniques, so it is associated with a spatial field, $\bs{o} =  \mathcal{O}(\bs{u}, \bs{f}, \bs{g})$, that is, $\bs{o} = \bs{o}(\bs{r})$, $\bs{r} \in \mathcal{R}$. As PGNNIV works at a discrete level (pixels, voxels, time frames...), $\mathbf{y}$ is a discrete representation of the spatial field $\bs{o}$. For instance, if $\mathcal{O}(\bs{u},\bs{f},\bs{g}) = \bs{u}$ and $\bs{u}$ is a $c$-dimensional spatial field that is discretized at $N$ points, then $\bs{Y} : \Omega \rightarrow \mathbb{R}^{c\times N}$ and $\mathbf{y} = (\bs{u}_j)_{j=1}^N$. Our goal is to represent this relation at a much lower dimensional latent space by means of an embedding
\begin{subequations}\label{eq:representation}
\begin{align} 
    \mathbf{z} &= \bs{\phi}(\mathbf{x}), \label{eq:representation_phi} \\
    \mathbf{y} &= \bs{\psi}(\mathbf{z}), \label{eq:representation_psi}
\end{align}
\end{subequations}
where $\bs{\phi} : \Omega \rightarrow (\mathbb{R}^{c})^n$ and $\bs{\psi} : (\mathbb{R}^{c})^n \rightarrow (\mathbb{R}^c)^N$. Here, $n \ll N$. Depending on the structure of $\bs{\phi}$ and $\bs{\psi}$, many approaches are possible, depending on the desired level of expressiveness and also on the knowledge about the experimental or computational set-up. The most straightforward idea is to assume a specific structure for $\bs{\psi}$ and therefore to learn the structure of $\phi$. This approach falls within the scope of spectral approximation theory \cite{boyd2001} (and in particular, Fourier analysis) and polynomial chaos expansion \cite{wiener1938}.

In addition to these \emph{a priori} approaches, a range of modern methods has emerged to focus on constructing the latent representation directly from empirical data \emph{a posteriori}. One the one hand, Proper Orthogonal Decomposition (POD) \cite{chatterjee2000introduction} is a data-driven approach that focuses on a particular choice of basis functions, which are constructed directly from the data. Its main advantage is its ability to capture the essential structures of the system while minimizing the number of modes or variations in the data. On the other hand, autoencoders have emerged as an alternative approach \cite{magri2022interpretability}. An autoencoder is a specific DNN architecture that learns a compact representation of the data, allowing for flexible and nonlinear approximations that can adapt to complex patterns in the data. In the next subsections, we summarize the different possible approaches from the perspective of the Eq. (\ref{eq:representation}).

\clearpage
\subsubsection{Spectral decomposition} \label{subsec::methods_spectral_decomposition}
Spectral decomposition is a method for representing functions as expansions in terms of orthogonal functions. Suppose we want to approximate a function $f : \Omega \rightarrow \mathbb{R}^M$ over some domain of interest as a finite sum in the variables separated form:
\begin{equation} \label{eq::spectral_approximation}
    f(x) \approx f_n(x) = \sum^{n}_{k=0} a_k \phi_k(x),
\end{equation}
where $a_n$ are the $n+1$ spectral coefficients with respect to the orthogonal basis $\mathcal{B} = \{\phi_k\}_{k=0}^\infty$; under the assumption that $\mathcal{B}$ forms a complete orthonormal system in a Hilbert space $\mathcal{H}$, the approximation converges to the original function in the norm of $\mathcal{H}$ as $n \to \infty$ \cite{katznelson2004introduction}. In the finite scenario, we can express $f(x) = f_n(x) + \varepsilon(x)$ where $\varepsilon$ is the approximation error. A good approximation is such that the error is minimized with respect to a certain distance (the theory of Hilbert spaces guarantees this approximation for the $L^2$ norm, by means of the Hilbert projection theorem \cite{rudin1973functional}).

The representation of this approximation is not unique. For instance, if the domain of $x$ is a bounded interval $x \in I \subset \mathbb{R}^n$, then the functions $\phi_k$, $k\in \mathbb{N}$ can be chosen as a Fourier basis, Legendre polynomials, Chebyshev polynomials, or other orthogonal polynomials. Each choice of the functions $\phi_i$ that forms a basis for some suitable class of functions $f$, leads to a different sequence of coefficients $a_k$, $k\in \mathbb{N}$.

For the sake of simplifying the notation, but without loss of generality, let us assume that $\bs{O}(\bs{u},\bs{f},\bs{g}) = \bs{u}$. Then, $\bs{u} = \bs{u}(\bs{r})$, $\bs{r} \in \mathcal{R}$. Therefore, it is possible to express
\begin{equation}\label{eq::spectral_decomposition_u}
    \bs{u}(\bs{r};\mathbf{x}) \approx \bs{u}_n(\bs{r};\mathbf{x}) = \sum_{i=0}^n u_i(\mathbf{x})\phi_i(\bs{r}).
\end{equation}
Hence, instead of learning the relation $\mathbf{x} \mapsto \bs{u}(\bs{r}),\;  \forall \bs{r} \in \mathcal{R}$, we learn the relation $\mathbf{x} \mapsto u_k, \forall k=1,\ldots,n.$ For the discrete case where the field $\bs{u}$ is represented by an array $\mathbf{u} = \bs{u}^{(N)} = (\bs{u}(\bs{r}_j))_{j=1}^N$, savings are evident if $n \ll N$.

One of the most widely used spectral basis is the Fourier basis. Let $\bs{u}(\bs{r}; \mathbf{x})$ be a square-integrable function defined over a compact domain. Then $\bs{u}$ can be represented using the Fourier transform as:
\begin{equation}\label{eq::fourier_transform}
    \bs{u}(\bs{r}; \mathbf{x}) = \sum_{k=0}^n u_k(\mathbf{x}) e^{2\pi i \bs{\xi}_k \cdot \bs{r}},
\end{equation}
where $\bs{\xi}_k$ are the wave vectors, and $u_k(\mathbf{x})$ are the Fourier coefficients given by:
\begin{equation}\label{eq::fourier_transform_coef}
    u_k(\mathbf{x}) = \int_\mathcal{R}\bs{u}(\bs{r}; \mathbf{x}) e^{-2\pi i \bs{\xi} \cdot \bs{r}} \, \mathrm{d}\bs{r}.
\end{equation}

This formulation allows $\bs{u}(\bs{r}; \mathbf{x})$ to be decomposed into a sum of harmonic modes. By selecting a finite number of modes, we can reconstruct an approximation of the original function. For practical applications involving discrete and finite data, the discrete Fourier Transform (DFT), or its optimized version, the Fast Fourier Transform (FFT) \cite{cooley1965fourier} can be used. The method will be coupled with the PGNNIV framework and its implementation is detailed in section \ref{subsec::predictive_component}.


\subsubsection{Proper Orthogonal Decomposition} \label{subsecc::methods_POD}

The POD is a numerical method that facilitates the reduction of the complexity of computer-intensive simulations \cite{chatterjee2000introduction}. The primary objective of POD is to extract dominant structures from a physical field represented by large amounts of high-dimensional data. This is achieved by applying the Singular Value Decomposition (SVD) to a matrix of samples. The resulting decomposition produces a set of orthonormal modes that optimally represent the data by minimizing the matrix 2-norm (spectral norm) of the difference between the original matrix $A$ and its reconstruction. The matrix 2-norm is defined as
\begin{equation}\label{eq:matricial_norm}
    \left|\left| \bs{A} \right|\right|_2 = \sqrt{\lambda_{\max}(A^*A)},
\end{equation}
where $A^*$ denotes the conjugate transpose of $A$, and $\lambda_{\max}(A^*A)$ is the largest eigenvalue of the positive semi-definite matrix $A^*A$.

Consider a system that can collect data in which measurements of $N$ state variables can be taken at $D$ experimental observations. The data can be arranged in a matrix $\bs{A} \in \mathbb{R}^{D \times N}$, known as snapshots matrix, where each column corresponds to a specific measurement and each row to a specific experiment. We now compute the SVD of the matrix $\bs{A}$: 
\begin{equation} \label{eq::SVD}
    \bs{A} = \bs{U} \bs{\Sigma} \bs{V}^{\intercal},
\end{equation}
where $\bs{V}^\intercal$ is the transpose of an $N \times N$ orthogonal matrix, whose rows represent the spatial modes of the system; $\bs{U}$ is an $D \times D$ orthogonal matrix, whose columns represent the components of the corresponding spatial mode along all the experiments; and $\bs{\Sigma}$ is an $D \times N$ diagonal matrix with all elements zero except along the diagonal. The diagonal elements $\sigma_i = \Sigma_{ii}$ represent the energy of each mode and consist of $r  =  \mathrm{rank}(\bs{A}) \leq \min(N, D)$ non-negative numbers, called the singular values of $\bs{A}$, arranged in decreasing order: $\sigma_1 \geq \sigma_2 \geq \cdots \geq \sigma_r \geq 0$. The rank of $\bs{A}$ is equal to the number of nonzero singular values it has. For any $k < r$, the matrix $\bs{\Sigma}_k$ obtained by setting $\sigma_{k+1} = \sigma_{k+2} = ... = \sigma_{r} = 0$ in $\bs{\Sigma}$ can be used to calculate an optimal rank $k$ approximation to $\bs{A}$, given by
\begin{equation} \label{eq::SVD_approximation}
    \bs{A} \approx \bs{A}_k = \bs{U} \bs{\Sigma}_k \bs{V}^\intercal,
\end{equation}
in terms of the spectral norm given by Eq.  (\ref{eq:matricial_norm}). We would replace $\bs{U}$ and $\bs{V}$ with the matrices of the first $k$ columns and replace $\bs{\Sigma}_k$ by its leading $k \times k$ principal minor, the submatrix consisting of $\bs{\Sigma}$'s first $k$ rows and $k$ first columns. 

To ensure that the chosen modes capture the majority of the system's dynamics while limiting the truncation error, the number of modes $k$ is chosen such that the error $\epsilon$ is minimized, thereby preserving most of the system's energy. Mathematically, this can be expressed as:
\begin{equation}
    \epsilon =  1 - \frac{\sum_{i=1}^k \sigma_i^2}{\sum_{i=1}^N \sigma_i^2},
\end{equation}
where $\sigma_i$ are the singular values ordered in decreasing magnitude, $N$ is the total number of modes, and $k$ the number of chosen modes, with $k < N$, and which determines the number of elements in the latent space.

Let us suppose we collect data of a physical system $\bs{u}(\bs{r}; \mathbf{x})$ in a matrix $\bs{A} = (\bs{u}_i(\mathbf{x}_j))_{i=1,\ldots,N,j=1,\ldots,D}$. Now we define $\bs{Q} = \bs{U} \bs{\Sigma}$ (or $\bs{Q} = \bs{U} \bs{\Sigma}_k$ in the reduced case). This decomposition allows us to separate our system into two components,
\begin{equation} \label{eq::POD_for_PGNNIV}
    \bs{A} = \bs{Q} \bs{V}^\intercal
\end{equation}
where matrix $\bs{Q}$, represents in its columns the behavior of a mode across all experiments indexed by $i$, scaled by the energy (singular value) associated with each particular mode; and matrix $\bs{V}^\intercal$ contains in its rows the basis of the orthogonal modes of the system. This formulation corresponds to the matrix representation of Eq.(\ref{eq:representation}), where multipliying by $\bs{Q}$ corresponds to Eq.(\ref{eq:representation_phi}) and by $\bs{V}^\intercal$ to Eq.(\ref{eq:representation_psi}). The method will be coupled with the PGNNIV framework and its implementation is detailed in section \ref{subsec::predictive_component}.

\subsubsection{Autoencoders} \label{subsec::methods_autoencoder}

An autoencoder \cite{magri2022interpretability} is a specific type of DNN architecture that approximates the identity operator and learns a new representation of the input data. The model consists of two main components: an encoder, which compresses the original data to a latent space; and a decoder, which maps the latent space back to the original space, providing an approximation of the input vector. Their architecture is known as a bottleneck. The model is characterized by a reduction in the number of neurons in each layer during the encoding phase, until the latent space (the layer with the fewest neurons) is reached. This structure is mirrored in the decoding phase, where the number of neurons is increased to reconstruct the original input at the output layer. 


We take inspiration of autoencoders architecture. In order to approximate a certain physical field given by $\bs{u}(\bs{r}; \mathbf{x})$ with a reduced-order representation $\tilde{\bs{u}}(\bs{r}; \mathbf{x})$, an autoencoder will map the original physical domain into a manifold, and then maps it back into the measurement physical space. In terms of neural networks, the approximation given by the spectral decomposition can be rewritten as an encoding model $\mathbf{z} = \bs{\phi}(\mathbf{x})$, and a decoding model $ \tilde{\bs{u}}(\bs{r}; \mathbf{x}) = \bs{\psi}(\bs{z}(\mathbf{x}))$, as represented in Eq. (\ref{eq:representation}). In terms of spectral decomposition of functions, this can be interpreted as:
\begin{equation}
    \tilde{\bs{u}}(\bs{r}, \mathbf{x}) = \sum_{k=0}^n u_k(\mathbf{x}) \bs{\Psi}_k  \left( u_k(\mathbf{x}), \bs{r} \right)
\end{equation}
where $u_n(\mathbf{x})$ are the components of $\tilde{\bs{u}}$ on the curvilinear basis $\mathcal{B} = \{\bs{\Psi}_n  \left( u_n(\mathbf{x}), \bs{r} \right)\}$ (in general nonlinear non-orthogonal), which approximates the output field nonlinearly. The objective of the autoencoder is to find this optimal basis $\bs{\Psi}$ that minimizes the approximation error. An autoencoder-like architecture is coupled with the PGNNIV framework and its implementation is detailed in section \ref{subsec::predictive_component}.

\clearpage
\subsection{Model complexity} \label{subsec::model_complexity}
The use of embedding representations for the predictive network has a direct impact on the model complexity. To evaluate the PGNNIV model complexity, we need to take into account its parametric learning space. For the baseline PGNNIV the number of parameters for both the predictive and explanatory network is:
\begin{subequations}\label{eq::hp_pgnniv}
    \begin{equation}\label{eq::hp_pgnniv_pred}
        P_{\text{pre}} = i \cdot h_0 + \sum^{L-2}_{k=0} h_k \cdot h_{k+1} + o \cdot h_{L-1} + \sum^{L-1}_{k=0} h_k + o,
    \end{equation}
    \begin{equation}\label{eq::hp_pgnniv_exp}
        P_{\text{exp}} = c_\text{in} n + h'_0 n + \sum^{L'-1}_{k=0} h'_k \cdot h'_{k+1} + h'_L n + \sum^{L'}_{k=0} h'_k + 2n + c_\text{out}n + c_\text{out},
    \end{equation}
\end{subequations}
where $h_k$ for $k = 0, \ldots, L$ represents the number of neurons in the $k$-th hidden layer in the predictive network and  $h'_k$ for $k = 0, \ldots, L'$ in the explanatory network, $o$ corresponds to the number of elements in the output layer, and $n$ is the number of convolutional $1 \times 1$ filters used to expand the single-neuron input of the explanatory network, and $c_\text{in}$ and $c_\text{out}$ are the number of channels in the input and the output of the network. Eq. (\ref{eq::hp_pgnniv_pred}) specifies the total number of trainable parameters associated with the predictive component of the network, and Eq. (\ref{eq::hp_pgnniv_exp}) defines the number of trainable parameters corresponding to the explanatory component of the network. The simplicity of this network is due to its nature of single-neuron input/output network. In both expressions weights and biases of the model are considered as trainable parameters.

Eqs. (\ref{eq::hp_pgnniv}) indicates how the model size (i.e. the number of trainable parameters) is influenced not only by the number of hidden layers and the number of neurons in each layer (including the latent space, which is a layer that has been given a meaning), but also by the input and output size. In other words, the model size is dependent on the discretisation of the data. Suppose that the output variable is a high resolution $(p,q)$-order tensor field defined at a mesh grid in dimension $d$. If $\Delta$ is the characteristic dimension of the problem and  $\delta$ is the characteristic resolution, then $o \sim \left(\frac{\Delta}{\delta}\right)^d d^{p+q}$ making explicit both the curse of dimensionality and the impact of the resolution.

One of the main goals of this work is to improve PGNNIV performance, by means of reducing the model complexity while preserving the model accuracy. One of the approaches involves separating its predictive component into two parts: the encoding network and the decoding network. This allows the decoding process to be replaced by a field reconstruction from the latent space (by means of one of the previously mentioned approaches). Therefore, it is useful to express Eq. (\ref{eq::hp_pgnniv_pred}) in a way that distinguishes between these two networks:
\begin{equation}\label{eq::hp_pgnniv_split}
    P_{\text{pre}} = 
    \underbrace{ i \cdot h_0 + \sum_{k=0}^{B-1} h_k \cdot h_{k+1} + \sum_{k=0}^{B}h_k }_{\text{Encoding network}} 
    \quad + \quad 
    \underbrace{ \sum_{k=B}^{L-1} h_k \cdot h_{k+1} + o \cdot h_N + \sum_{k=B}^{L-1} h_{k+1} + o }_{\text{Decoding network}},
\end{equation}
where $B$ is the number of layers up to the bottleneck. Note that the output size does not appear in the encoding block, thus not affecting the number of parameters. Therefore, the strategy bypasses both the resolution scale issue and the curse of dimensionality.

For illustration purposes, let us consider the following architecture for the PGNNIV. In the predictive network, the inputs, outputs, and latent space are treated as free parameters. Thus, the architecture of its internal layers is defined as $h_{\text{pred}} = [20, 10, n, 10, 20]$. In the case of the explanatory network, there are no variable parameters. This network has a 1-channel input and 1-channel output, so $c_\text{in} = c_\text{cout} = 1$, it employs $n_{\text{filters}} = 5$ convolutional $1 \times 1$ filters, and the number of neurons in the explanatory network, denoted as $h_{\text{exp}}$, is set to $10$, as it consists of a single-layer neuron. Let us suppose that the output $\mathbf{y}$ is a discretized 2D solution field on a mesh of size $N_x \times N_y$ and the input variable $\mathbf{x}$ is the collection of Neumann and Dirichlet boundary conditions. Then the input size is $i = 4N_x + 4N_y$, that is the (maximal) number of boundary conditions prescribed and the output size is $o = N_x N_y$, which is the total number of pixels in the output image. Substituting these values into Eq. (\ref{eq::hp_pgnniv_pred}) and Eq. (\ref{eq::hp_pgnniv_exp}) yields:

\begin{subequations}
\begin{equation}\label{eq::hp_pgnniv_pred_numeric}
    P_{\text{pre}} = P_{\text{encoding}} + P_{\text{decoding}} = \\ \underbrace{ 80(N_x + N_y) + 230 + 11 \ n }_{\text{Encoding network}}  + \underbrace{ 230 + 10 \ n + 21N_x N_y }_{\text{Decoding network}} 
\end{equation}
\begin{equation}\label{eq::hp_pgnniv_exp_numeric}
    P_{\text{exp}} = 161,
\end{equation}
\end{subequations}
Therefore, the explanatory network remains constant in complexity, while the predictive network depends on the data resolution and the size of the latent space. If the decoding network is removed and replaced by one of the methods described in subsection \ref{subsec::latent_space_embedding}, the component requiring the majority of the parameters is replaced by a single-step operation, such as reconstructing the solution from a fixed basis applying a matrix multiplication or running a forward pass of the pretrained decoder. In all cases, this operation is fixed and does not require parameter fitting. Notably, the decoding complexity term, which scales with $N_x N_y$ in two dimensions, often dominates due to the cost of reconstructing over the full spatial grid. Indeed, if $N_x = N_y = m$, then $P_\mathrm{pre} = \mathcal{O}(m^2)$ for the full network and $P_\mathrm{pre} = \mathcal{O}(m)$ for the one with the latent space embedding.

\begin{figure}[h!]
    \centering
    \includegraphics[]{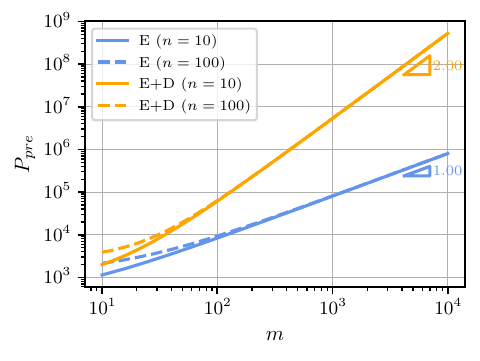}
    \caption{\textbf{Model scalability.} Evolution of the number of trainable parameters in the predictive network of a baseline PGNNIV model, as a function of data resolution and latent space size, depending on the situation: if the model has encoder and decoder (E + D) or if it is the encoder and other non-trainable operation (E) (as a matrix multiplication or inference step of a pretrained model).}
    \label{fig::HP_evolution}
\end{figure}

This effect is illustrated in Fig. \ref{fig::HP_evolution}, where the evolution of the number of trainable parameters of the encoding and decoding strcutures with respect to the resolution and latent space size are shown.

\subsection{Isolating experimental set-up and material behavior}

Let us consider an experimental set-up $\mathcal{E}$ that is able to generate pairs of data $(\mathbf{x},\mathbf{y})$. In some situations, it is possible to control the input variables $\mathbf{x}$ and to measure the variable $\mathbf{y}$ associated with each input value. However, in the context of Internet of Things (IoT) and massive sensor monitoring, we sometimes are interested in learning the relation $\bs{Y}$ for a given sample $\mathcal{S}$ without being able to control the values of the variable $\mathbf{x}$. Thus, the structure of a given dataset $\mathcal{D} = \{(\mathbf{x}^i,\mathbf{y}^i)\}_{i=1}^D$ depends both on the experimental set-up $\mathcal{E}$ and on the particular specimen $\mathcal{S}$.

To understand this statement, let us think of two different homogeneous and isotropic elastic materials, both assumed to be linear but with different elastic parameters $E$ and $G$, that are tested using three different experimental set-ups:  compression test, shear test and torsion test. For each of these experimental set-ups, the input and output variables are different:
\begin{itemize}
    \item $\mathrm{x} = F$ the compression force and $\mathrm{y} = \Delta$ the longitudinal displacement for the compression test. Under the assumption of linear isotropic elasticity, we know that $\mathrm{y} = E \mathrm{x}$.
    \item $\mathrm{x} = T$ the shear force and $\mathrm{y} = \theta$ the angular strain for the shear test. In that case $\mathrm{y} = G \mathrm{x}$.
    \item $\mathrm{x} = \tau$ the torque and $\mathrm{y} = \alpha$ the rotation angle. In that situation, $\mathrm{y} = \frac{GJ}{L}\mathrm{x}$.
\end{itemize}
In addition, the experimental data $\mathcal{D} = \{(\mathbf{x}^i,\mathbf{y}^i)\}_{i=1}^D$ also depends on the specimen considered $\mathcal{S}$. Thus, the set of all possible (homogeneous and isotropic linear elastic) specimens is a manifold $\mathcal{M}_\mathcal{S}$ of dimension $d_\mathcal{S} = 2$ (the parameters $E$ and $G$) and the manifold associated with the data generation process $\mathcal{M}_\mathcal{D}$ has dimension $d_\mathcal{D} = 1$, although it is different for each of the three different tests considered. It is clear that, in the general situation, we could have different values for $d_\mathcal{S}$ and $d_\mathcal{D}$. For instance, in a biaxial test $d_\mathcal{D} = 2$. It is clear that a good experimental set-up is one in which the distance between $\mathcal{M}_\mathcal{S}$ and $\mathcal{M}_\mathcal{D}$ is as small as possible.

The idea of isolating these two physical components (the one related to the sampling process and the one related with the material behavior)  may be envisaged using latent space embedding. Indeed, the information about the experiment sampling variability is encoded in the predictive network $\mathsf{Y}$, via the values of the latent variable $\mathbf{z}$, whereas the information about the material behavior is encoded in the explanatory network $\mathsf{H}$. For instance, for a biaxial test, regardless of how the data acquisition is performed, a predictive network with a latent space of dimension $n = 2$ should be able to capture the $\mathbf{x} \mapsto \mathbf{y}$ relationship. In addition, the explanatory network should be able to unveil the relation between the pairs strains - stresses. If the set-up is then changed to a shear test, the map $\mathbf{x} \mapsto \mathbf{y}$ is completely different, whereas the relation between the pairs strains - stresses remains unchanged.

This idea may be exploited easily under the PGNNIV framework. In addition to designing the predictive network in a more specific way, the fact that PGNNIVs isolate the variability dependent on the type of experiment from the variability dependent on the behavior of the material leverages the use of Transfer Learning in the following sense. If we have correctly learned the embedding manifold for a given experimental set-up, for instance using an autoencoder, there is no need of training the encoder again when trying to learn the behavior of a different material. If the architecture of the PGNNIV is described by the three components $(\bs{\phi},\bs{\psi},\mathsf{H})$, the encoder $\phi$ has to be trained only once and for all for a particular set-up, and then only the relation between the latent variable $\mathbf{z}$ and the outcome $\mathbf{y}$ has to be adapted for the new material, in addition to the explanatory network $\mathsf{H}$, of much smaller size and then less computationally demanding.

The procedure in a general context for an experimental set-up is as follows:
\begin{enumerate}
    \item Given an experimental set-up $\mathcal{E}$ with a particular specimen $\mathcal{S}$, we generate the data $\mathcal{D}$ depending both on $\mathcal{E}$ and $\mathcal{S}$.
    \item \textbf{Training step.} We train the PGNNIV $(\bs{\phi},\bs{\psi},\mathsf{H})$ using the dataset $\mathcal{D}$.
    \item \textbf{Material characterization.} The state equation is characterized by the PGNNIV component $\mathsf{H}$.
    \item \textbf{New specimen data generation.} For a new specimen $\mathcal{S}'$ that is tested using the same experimental set-up $\mathcal{E}$, we generate new data $\mathcal{D}'$.
    \item \textbf{New training step.} We train the PGNNIV components $(\bs{\psi}',\mathsf{H}')$, while keeping frozen the component $\bs{\phi}$, using the dataset $\mathcal{D}'$.
    \item \textbf{New material characterization.} The state equation of the new specimen is characterized by the PGNNIV component $\mathsf{H}'$.
\end{enumerate}
The previous procedure may be repeated for many different specimens.

\section{Methods} \label{secc::methods}

\subsection{PGNNIV structure} 

To illustrate the use of PGNNIV in continuum physical problems, as well as the improvements presented in this work, let us suppose the following PDE corresponding to a diffusion problem, as was done in \cite{munoz2025application}:
\begin{equation}
    -\bs{\nabla} \cdot (\bs{K} \bs{\nabla} u) = f,
    \label{eq:diffusion_PDE}
\end{equation} 
where $u$ is the solution field and $f$ is the source term. Eq. (\ref{eq:diffusion_PDE}) is the combination of two different laws: a fundamental principle, as it is energy conservation $\bs{\nabla} \cdot \bs{q} = f$, where $\bs{q}$ is the flux vector and $f$ is the source term; and a constitutive functional equation $\bs{q} = -\bs{K} \bs{\nabla} u$, relating the flux variable $\bs{q}$ that plays the role of internal state field (non-measurable if no additional assumption is made), with the essential field $u$, being $\bs{K}$ is the diffusion tensor. For general nonlinear problems, the tensor $\bs{K}$ may be dependent (in a functional sense) on the field $u$.

With these assumptions, Eq. (\ref{eq:diffusion_PDE}) is split in:
\begin{subequations}\label{eq:fundamental_principle_constitutive_equation}
    \begin{align}
        \bs{\nabla} \cdot \bs{q} &= f,\label{eq:fundamental_principle}  \\
        \bs{q} &= -\bs{K} \bs{\nabla} u, \label{eq:constitutive_equation} 
    \end{align}
\end{subequations}
together with appropriate boundary conditions. Here, the fundamental principle and the state equation are expressed in functional form. This is bypassed by using any common discretization technique so the values of $u$, $\bs{q}$ and $\bs{K}$ are replaced by the corresponding nodal values or by the approximation values, $\bs{u}^{(N)}$, $\bs{q}^{(N)}$ and $\bs{K}^{(N)}$. Of course, when solving Eqs. (\ref{eq:fundamental_principle_constitutive_equation}), or their corresponding discrete version, proper parametric boundary conditions must be supplied.

If the problem is now formulated within the PGNNIV framework, these latter boundary values, together with $\bs{f}^{(N)}$, are the natural inputs of the problem, being $\bs{u}^{(N)}$ the output ones. It is possible to predict the value of the fields $u$, $\bs{q}$ and $\bs{K}$ given the fields $f$ and $g$ by using the approach such that the nodal values of $u$ (that is $\bs{u}^{(N)}$) are learned from a sufficiently and varied dataset of input-output pairs. However, in this work, we propose the reconstruction of the nodal values $\bs{u}^{(N)}$ from the latent variable $\mathbf{z} \in \mathbb{R}^n$.

Next, we detail the architecture of the two PGNNIV components (the predictive and the explanatory network) as well as the physically inspired regularization procedure. For simplicity, we particularize Eqs. (\ref{eq:fundamental_principle_constitutive_equation}) for 2D problems.
\subsubsection{Predictive component} \label{subsec::predictive_component}

The predictive network is based on a Multilayer Perceptron (MLP) architecture with a bottleneck structure, as described in section \ref{subsec::methods_autoencoder}. Although its design remains of a conventional autoencoder, due to the presence of a low-dimensional latent space, the network's role is significantly different. In contrast to a traditional autoencoder, whose aim is to learn an identity mapping reconstruct the input, the predictive network is trained to predict the output data, different from input data. In this sense, it behaves as a standard feedforward MLP, where the bottleneck serves to encourage the extraction of relevant features. This design enables the model to learn latent representations that are optimized for the prediction task. Let $\bs{u}^{(N)}(\bs{r}; \mathbf{x}) = \bs{u}(r_k, r_l; \mathbf{x})$ the discrete representation of the solution of Eq. (\ref{eq:diffusion_PDE}), as a two-dimensional field sampled on a uniform grid with $k = 0, \dots, N_x - 1$ and $l = 0, \dots, N_y - 1$, where $N= N_x \times N_y$  is the total number of degrees of freedom. In a similar manner, let $\bs{q}_x^{(N)}(\bs{r}; \mathbf{x}) = \bs{q}_x(r_k, r_l; \mathbf{x})$ and $\bs{q}_y^{(N)}(\bs{r}; \mathbf{x}) = \bs{q_x}(r_k, r_l; \mathbf{x})$ the internal flux, which can only be measured at the contour. The boundary conditions $\bs{g}^{(N)}(r_k, r_l; \mathbf{x})$ of the problem are these contour values, both of the solution $\bs{u}$ and of the flows $\bs{q}_x$ and $\bs{q}_y$. These boundary conditions are inputs $\mathbf{x}$ of the predictive model, while the output data $\mathbf{y}$ are the complete (discretized) solution field $\bs{u}^{(N)}$.

\begin{figure}[h!]
    \centering
    \includegraphics[width=1\linewidth]{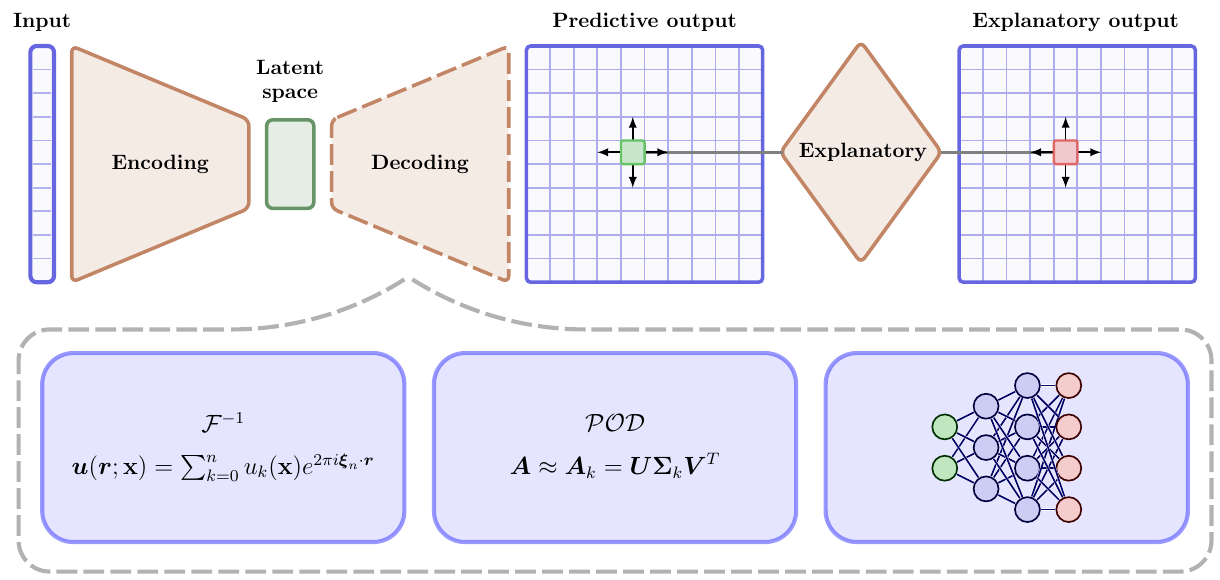}
    \caption{\textbf{Assembly of the different decoding strategies within the general PGNNIV model.} PGNNIV where the decoder is substituted by the use of a Fourier basis (left), by computing the POD (center), and by a pretrained autoencoder (right), .}
    \label{fig:embedding_representations_scheme}
\end{figure}

In addition to the baseline predictive network presented in \cite{munoz2025application}, we will implement three different methods to decode these learned latent representations and map them into the original space, thereby obtaining the predicted solution: a spectral decomposition in the Fourier basis, the Proper Orthogonal Decomposition and the use of the decoder of a previously trained  autoencoder. In Fig. \ref{fig:embedding_representations_scheme} it is represented a graphical scheme of the three embedding methods.

\paragraph{Fourier spectral basis.} On the one hand, in the Fourier-based method, the decoder of the predictive network is removed. Instead, it will be replaced by the multiplication of the latent space by the Fourier basis. In other words, the inverse Fourier transform of the latent space will be computed. This design enables the network's direct learning of the Fourier coefficients, which are represented within the network's latent space, instead of computing them from the data. The inverse Fourier Transform reconstructs the field as
\begin{equation}
    \bs{u}(r_p, r_q; \mathbf{x}) = \sum_{k_1=0}^{N_x-1} \sum_{k_2=0}^{N_y-1} u_{k_1k_2}(\mathbf{x}) \, e^{2\pi i \left( \frac{k_1r_p}{n_x} + \frac{k_2r_q}{n_y} \right)}
\end{equation}
from the Fourier coefficients $u_{k_1k_2}(\mathbf{x})$. Thus, with the implementation of this method, once the predictive network has reached an appropriate solution, it will have learned an approximation of the coefficients $\hat{u}_{k_1k_2}(\mathbf{x}) \approx u_{k_1k_2}(\mathbf{x})$, directly in its latent representation. Here, the size of the latent space is $n = n_x \times n_y$ where $n_x$ and $n_y$ are the number of selected nodes in the $x$ and $y$ directions.

It is important to note that the strategy involving spectral approximation does not require more computation time beyond the one of the PGNNIV training process.

\paragraph{Orthogonal basis obained via POD.} On the other hand, in the POD-based method, the decoder is replaced by a linear reconstruction using a precomputed set of orthonormal basis functions derived from the training data. To reduce the dimensionality of the problem, the set of different solutions of the two-dimensional field $\bs{u}(r_k, r_l; \mathbf{x})$ is arranged in the snapshot matrix, where each row corresponds to a different experiment and each column corresponds to a spatial measurement. The three matrices obtained are: $\bs{U}$, that contains the mode activations (their behavior along all the experiments), $\boldsymbol{\Sigma}$, that contains the energy associated with each mode, and $\bs{V}^\intercal$, that contains the  orthonormal spatial modes of the system. The matrix $\bs{V}^\intercal$ will be kept in order to project the learned latent space into the original physical space and obtain the solution of the problem. This is equivalent to compute Eq. (\ref{eq::POD_for_PGNNIV}), where once the predictive network has reached an appropriate solution, it will have learned an approximation of the matrix $\hat{\bs{Q}} \approx \bs{Q}$.

For the POD strategy, the extra computational effort needed, in addition to the PGNNIV training process, is the one of a SVD computation for the dataset snapshot matrix $\bs{A}$.

\paragraph{Local nonlinear basis obtained via autoencoders.} Finally, we consider the autoencoder-based approach. In this method, an autoencoder is trained using the field $\bs{u}^{(kl)}(\mathbf{x}) = u(r_k, r_l; \mathbf{x})$ as input and output. The objective of the autoencoder is to minimize the difference between the input and the output. The loss function is:
\begin{equation}
    \mathcal{L}_{\text{autoencoder}} = \sum_{i=1}^D\| \bs{u}^{(kl)}(\mathbf{x}^i) - \hat{\bs{u}}^{(kl)}(\mathbf{x}^i)\|
\end{equation}
where $\hat{\bs{u}}^{(kl)}(\mathbf{x})$ is the output of the model corresponding to the configuration associated with $\mathbf{x}$. Thus, the encoder will map the input into a compact latent representation, while the decoder will reconstruct the original field. Once autoencoder training is complete, the decoder is extracted and incorporated into the predictive model as a fixed component, replacing the trainable decoder. 
Since the decoder is pretrained and the rest of the predictive network is not, it is kept frozen during the training of the PGNNIV: its parameters will not be subject to further optimization. Consequently, the training process focuses exclusively on learning the mapping from the input parameters to the latent representation of the field. The decoder thus functions as a nonlinear map, reconstructing the full field from the learned latent space.

The strategy involving the decoder pre-training step is the more computationally expensive as it requires, in addition to the PGNNIV training process, to train a complete autoencoder. However, this task can be performed only once for a given experimental configuration, and then used for different specimen characterizations, as will be discussed later.

\subsubsection{Explanatory component}

The explanatory network is a convolution-inspired architecture that implements a moving MLP (mMLP) \cite{ayensa2025predicting} to process input images and capture local nonlinearities in the data. It starts with a sliding kernel that computes a $1 \times 1$ convolution, expanding each input pixel into a higher-dimensional representation of $n_c$ channel feature space, acting as a channel-wise transformation. The result is then transformed so that each $n_c$ transformed pixels becomes a separate input sample to an MLP consisting of one hidden layer, the output being also one layer of $n_c$ elements. Finally, a second $1\times 1$ convolution reduces the feature dimension back to a single channel output. This process is performed for each of the pixels in each of the images. This architecture emulates the locality and translation invariance of traditional convolutional layers. However, it replaces the linear kernel operation with a shared neural network, allowing more expressiveness in capturing non-linear patterns at the pixel level while preserving the spatial structure of the input. 

The explanatory network architecture is the same (although its parameters are trainable), regardless of any modifications to the predictive network. Its structure and parameters are held constant, ensuring that the model remains agnostic to the underlying constitutive law and does not incorporate any prior knowledge before training.

\subsubsection{Coupling both networks via loss regularization}
To link Physics-based and ML problems, we must appropriately select measurable input-output variables (as they are boundary conditions and field solutions), to embed the physical constraints and constitutive laws into the network via Prescribed Internal Layers (PILs), and to approximate the partially known constitutive relationships with a learnable sub-network. Further details on the PGNNIV methodology can be found in \cite{ayensa2021prediction, ayensa2025predicting}. This three steps must be carefully balanced and maintained consistently. These constraints can be derived from universal balance laws or can incorporate additional knowledge about the system. Consistently with the standard approach in NNs, they must be used to minimize a cost function, that will facilitate the learning of both the output variables and the constitutive model:
\begin{equation}\label{eq:cost_function_general}
    \mathcal{L} = \mathcal{L}_{\text{data}} + \mathcal{L}_{\text{physics}} + \mathcal{L}_{\text{model}}
\end{equation}
where in our case, $\mathcal{L}_{\text{physics}}$ represents the physics associated with energy conservation, Eq. (\ref{eq:fundamental_principle}) and no extra term $\mathcal{L}_{\text{model}}$ is added to the loss.

\subsection{Data generation}\label{subsec::data_generation}
In order to test the methodology, we consider the solution of two different nonlinear isotropic problems with two different relations $\bs{K} = K(u)\bs{I}$, where in the first problem (Material 1) $K_1(u) = u(1-u)$, and in the second problem (Material 2) $K_2(u) = \frac{1}{1 + \mathrm{e}^{-5(u-2)}}$. For both problems, we generate a synthetic dataset using a ground truth solution that is obtained using the method of manufractured solutions \cite{roache2002code}, similarly as it was done in previous works \cite{munoz2025application}. The analytical expression of the solution field $u$, the flux vector $\bs{q}$, the diffusion field $K$ and the source term $f$ are reported in Table \ref{tab:material_expressions} for the two diffusion fields considered.

We define the following expressions of the field $u$, diffusivity $K(u)$, flux $\bs{q}$, and source term of the equation $f$, that satisfy identically Eq. (\ref{eq:fundamental_principle_constitutive_equation}) for the two materials mentioned above.

\begin{table}[h!]
    \centering
    \renewcommand{\arraystretch}{1.6}
    \begin{tabular}{ccc}
        \hline
        \textbf{Field} & \textbf{Material 1} & \textbf{Material 2} \\
        \hline
        \(u(x,y)\) & 
        \(\sqrt{a + bx + cy}\) & 
        \(a + bx + cy\) \\

      \(K(x,y)\) & 
        \(\left(1 - \sqrt{a + bx + cy}\right) \sqrt{a + bx + cy}\) & 
        \(\frac{1}{1 + e^{-5(a + bx + cy - 2)}}\) \\

        \(\bs{q}(x,y)\) & 
        \(\begin{pmatrix} \frac{b}{2} (1 - \sqrt{a + bx + cy}) \\[6pt] \frac{c}{2} (1 - \sqrt{a + bx + cy}) \end{pmatrix}\) & 
        \(\frac{1}{1 + e^{-5(a + bx + cy - 2)}} \begin{pmatrix} b \\ c \end{pmatrix}\) \\

        \(f(x,y)\) & 
        \(\frac{1}{4\sqrt{a + bx + cy}} (b^2 + c^2)\) & 
        \(\frac{5 e^{-5(a + bx + cy - 2)}}{\left[ 1 + e^{-5(a + bx + cy - 2)} \right]^2} (b^2 + c^2)\) \\
        \hline
    \end{tabular}
    \vspace{0.5em}
    \caption{\textbf{Analytical solutions used for dataset generation and error evaluations.} Field $u$, diffusivity $K$, flux $\bs{q}$, and source term $f$ of the equation for Material 1 and Material 2.}
    \label{tab:material_expressions}
\end{table}

\subsubsection{Input and output variables: spatial discretization.}
$D$ profiles of the field $u(x, y; a, b, c)$ were synthetically generated, by sampling solutions of a parametric family, obtained for different values of $a, b, c \sim \mathcal{U}[0, 1]$ randomly and independently generated. Then, from this solutions, boundary conditions are assigned as: $g_1 = u(x = 0, y)$, $g_2 = u(x = 1, y)$, $g_3 = u(x, y = 0)$ and $g_4 = u(x, y = 1)$. The same procedure is followed for the flux $q_x(x, y), q_y(x, y)$, where $q_1 = q(x = 0, y)$, $q_2 = q(x = 1, y)$, $q_3 = q(x, y = 0)$ and $q_4 = q(x, y = 1)$. These boundary conditions (solution field and flux) were considered also as input variables to ensure that the PGNNIV is associated with a well-posed problem \cite{munoz2025application}. The values of the field $u$ were particularized at $N_x \times N_y$ points where $N_x = N_y = m$, $x_i = y_i = \frac{i - 1}{m - 1}$, $i = 1, \ldots, m$ so the output variables correspond therefore to the nodal values $ u(x_i, y_j)$, $i,j = 1, \ldots, m$. Here we select $m=10$.

\paragraph{Data size and noise level.}
In order to carry out the experiments and evaluate the accuracy of the models under varying data conditions, a range of datasets have been generated. A variety of sample sizes $D=\{10, 100, 1000\}$ have been generated and, for each dataset, additive noise was introduced. The noise level is generated as follows: for each piece of data $i$, we define $B_{u,k} = \max\{u(x_i,y_j;\mathbf{x}^k) \, | \,  i,j=1,\ldots,m\}$ and we generate the noisy field by adding a noise $\varepsilon_k \sim \mathcal{N}(0, \sigma_k^2)$, where the standard deviation $\sigma_k$ is defined as $\sigma_k = \mu B_{u,k}$ with $\mu \in \{0.00, 0.01, 0.05\}$ is a scalar parameter that controls the amplitude of the noise. We do analogously for the fields $q_x$ and $q_y$.

\subsection{Detailed architecture}
The architecture of the predictive network consists of a fully connected five-layered neural network with the following neurons in its hidden layers: $h_{\text{pred}} = [20, 10, n, 10, 20]$, where $n$ is the latent space size. In the case of the embedding models, as it has been mentioned before, the decoder is substituted by a non-trainable operator, so the network will be reduced to a three-layered neural network with hidden layers $h_{\text{pred}} = [20, 10, n]$;   With the aim of capturing the most important modes of variation within the input data, the size of the latent space will be systematically varied. Specifically, three different models are evaluated, with $n \in \{5, 10, 50\}$ neurons in the latent space layer. With the mentioned discretization in which $m = 10$, the input size is given by $i = 4N_x + 4N_y = 8m = 80$, and the output size is given by $o = N_x \times N_y = m^2 = 100$. In the case of the explanatory network, none of the parameters is varied:  the network takes a single-channel input and produces a single-channel output, so $c_{\text{in}} = c_{\text{out}} = 1$, and considers $n_{\text{filters}} = 5$ convolutional filters of size $1 \times 1$, and a single hidden layer with $h_{\text{exp}} = 10$ neurons.

\paragraph{Training process.}
To ensure robust model evaluation and avoid overfitting, the dataset was divided into separate subsets for validation and testing. Specifically, 20\% of the total data was saved for post-training validation. Of the remaining part, a 30\% was reserved for in-training testing. In the case of the autoencoder, a more nuanced data splitting strategy was employed. As in the previous case, 20\% of the total data was saved for post-training validation. However, the remaining dataset, was segmented such that 50\% was allocated for training the autoencoder, while the remaining 50\% was reserved for training the full PGNNIV with the pretrained autoencoder. For both of the trainings an additional split was performed: 70\% for training and 30\% for internal testing during the training phases.

To ensure the effective training of all models for different datasets, a standardized training process has been implemented. As initialization of weights and biases of the models, the default PyTorch initializer has been used, which can be found in the official documentation \cite{pytorch2025docs}. As optimizer, we use the Adaptive Moment Estimation (Adam) method \cite{Kingma2014AdamAM}. A two-phase training scheme was computed: an initial training phase of $10^5$ epochs with a learning rate of $ \mathrm{lr} = 3 \times 10^{-3}$, followed by a second phase of $5\times 10^4$ additional epochs with a learning rate of $\mathrm{lr} = 3 \times 10^{-4}$. This strategy allows the models to converge quickly in the early stages while enabling fine-tuning in later stages through more conservative updates. 

A cost function is defined following the structure of Eq. (\ref{eq:cost_function_general}), and considered as the mean squared error (MSE) along the dataset, both for the error and penalty terms, and understood as the sum of the squares of all its components. For the diffusion problem considered, the penalty term equation is 
\begin{equation}\label{eq:penalties_sumatory_general_expression}
    \mathcal{L} = c_0\text{MSE}(e) + \sum_{i=1}^3 c_i \text{MSE}(\pi_i)
\end{equation}
where the different terms are the following: $\bs{e}$ is the prediction error between the output of the network $\hat{\mathbf{y}} = \mathsf{Y}(\mathbf{x})$ and the real solution $\mathbf{y} = \bs{u}$; $\bs{\pi}_1$ is difference between the contours of the predicted solution $\hat{\bs{u}}$ and the real contours of the solution ${\bs{u}}$; $\bs{\pi}_2$ is the difference between the contours of the predicted flows $\hat{\bs{q}}$, obtained by computing the Eq. (\ref{eq:constitutive_equation}) with the model estimations, and the measured ones; and finally, $\bs{\pi}_3$ is the residual of the Eq. (\ref{eq:fundamental_principle}) with the estimation of $\hat{\bs K}$ obtained by the explanatory network, using the predicted flow field $\hat{\bs{q}}$. Regularization is applied through penalties, which values are $c_0= 10^{7}$, $c_1 =10^{4}$, $c_2 = 10^{3}$, and $c_3 = 10^{5}$.

To ensure the reproducibility of all experiments, and allow fair comparisons, a fixed random seed was set at all relevant levels of the software stack, including Python’s \texttt{random} module and PyTorch’s CPU and GPU random number generators. Furthermore, deterministic behavior was enforced by configuring the CUDA backend to disable benchmarking and enable deterministic algorithms. This setup helps mitigate nondeterminism introduced by hardware and parallel computation, ensuring that identical runs yield consistent results.

\subsection{Performance evaluation}
To evaluate the predictive capacity of the PGNNIV, we evaluate the error in the prediction of the solution $\bs{u}$. To do so, we define the following prediction error
\begin{equation} \label{eq:predictive_error}
    \epsilon_r^{\text{pred}} = \frac{\sqrt{\int_{0}^{1} \int_{0}^{1}(\hat{u}(x, y) - u(x, y)^2 )\, \mathrm{d}x \mathrm{d}y}}{\sqrt{\int_{0}^{1} \int_{0}^{1}u(x, y)^2\, \mathrm{d}x \mathrm{d}y}},
\end{equation}
where $\hat{u}$ and $u$ are the predicted and real solution of the equation, respectively.

To evaluate the explanatory capacity of the PGNNIV, we evaluate the error in the discovery of the constitutive relation $K = K(u)$, that is, the ability of the method to unravel the relationship between the solution field $u$ and the diffusion coefficient $K$. To do so, we define the following explanatory error
\begin{equation} \label{eq:explanatory_error}
    \epsilon_r^{\text{exp}} = \frac{\sqrt{\int_{u_{\mathrm{min}}}^{u_{\mathrm{max}}} (\hat{K}(u) - K(u))^2 \, \mathrm{d}u}}{\sqrt{\int_{u_{\mathrm{min}}}^{u_{\mathrm{max}}}  K(u)^2 \, \mathrm{d}u}},
\end{equation}
where $\hat{K}(u)$ and ${K}(u)$ are the predicted and real diffusivity, respectively and $u_\mathrm{max}$ and $u_\mathrm{min}$ are the minimum and maximum values achieved by the solution field $u$.

\section{Results} \label{secc::results}

\subsection{Numerical experiments} \label{subsec::numerical_experiments}
For each of the different embeddings, a range of configurations is explored. These include varying the number of data samples $D$, the level of noise  $\mu$, and the size of the latent space $n$. This allows to systematically evaluate how these different parameters influence the capacity of the different approaches to learn meaningful representations of the data, maintain robustness under noisy conditions, and how it affects their explanatory capacity.

\subsubsection{Predictive and explanatory capacity}
\begin{figure}[h]
    \centering
    \begin{subfigure}[t]{0.48\linewidth}
        \centering
        \includegraphics[width=1\linewidth]{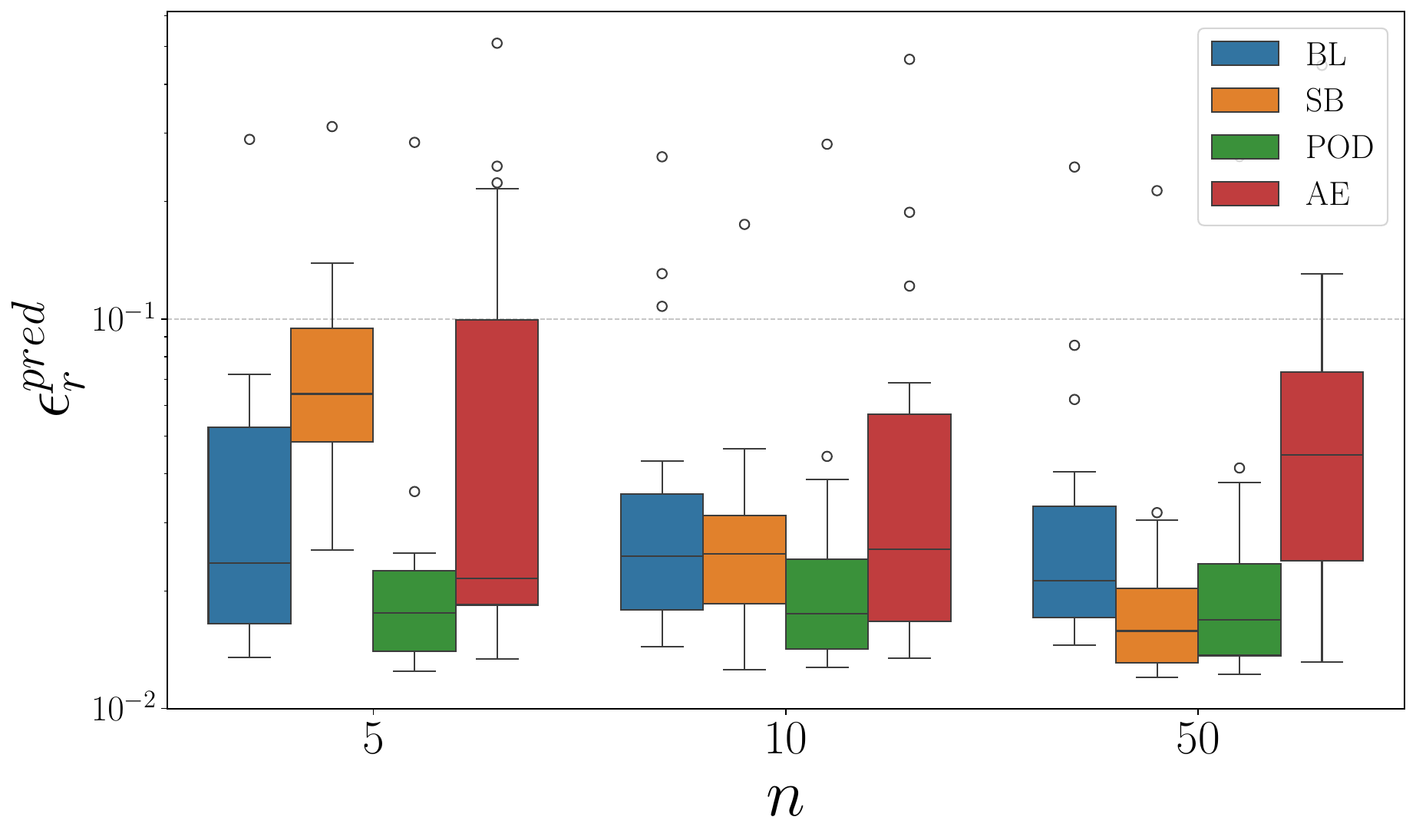}
        \caption{$\mu = 0.01$}
        \label{fig:pred_error_models_N100R1}
    \end{subfigure}
    \quad
    \begin{subfigure}[t]{0.48\linewidth}
        \centering
        \includegraphics[width=1\linewidth]{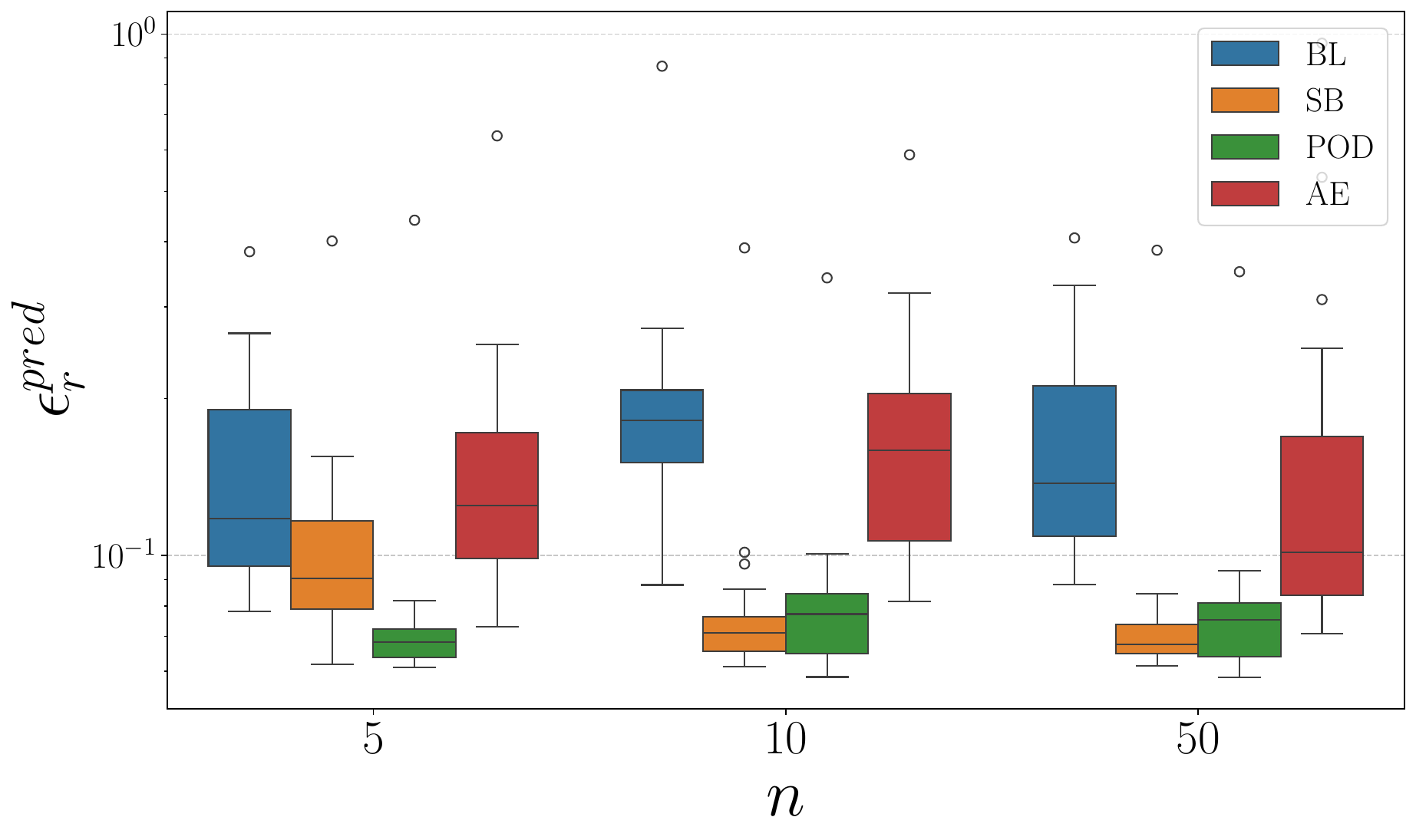}
        \caption{$\mu = 0.05$}
        \label{fig:pred_error_models_N100R5}
    \end{subfigure}
    \caption{\textbf{Predictive capacity for small dataset size.} Boxplots of the predictive error of the different approaches for $D=100$. BL: Baseline, SB: Spectral Basis, POD: Proper Orthogonal Decomposition, AE: Autoencoder.}
    \label{fig:pred_error_N100}
\end{figure}

\begin{figure}[h]
    \centering
    \begin{subfigure}[t]{0.48\linewidth}
        \centering
        \includegraphics[width=1\linewidth]{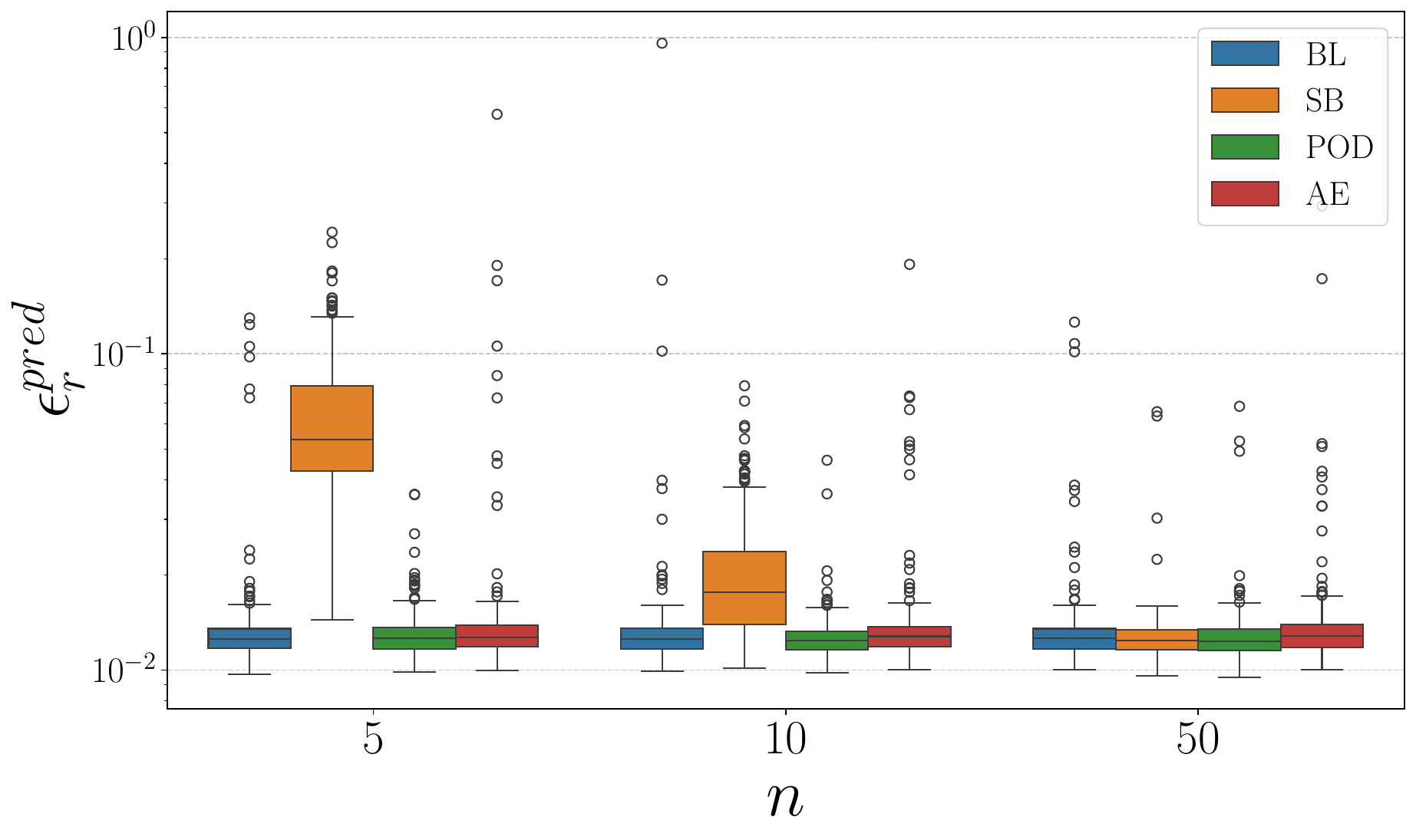}
        \caption{$\mu = 0.01$}
        \label{fig:pred_error_models_N1000R1}
    \end{subfigure}
    \quad \
    \begin{subfigure}[t]{0.48\linewidth}
        \centering
        \includegraphics[width=1\linewidth]{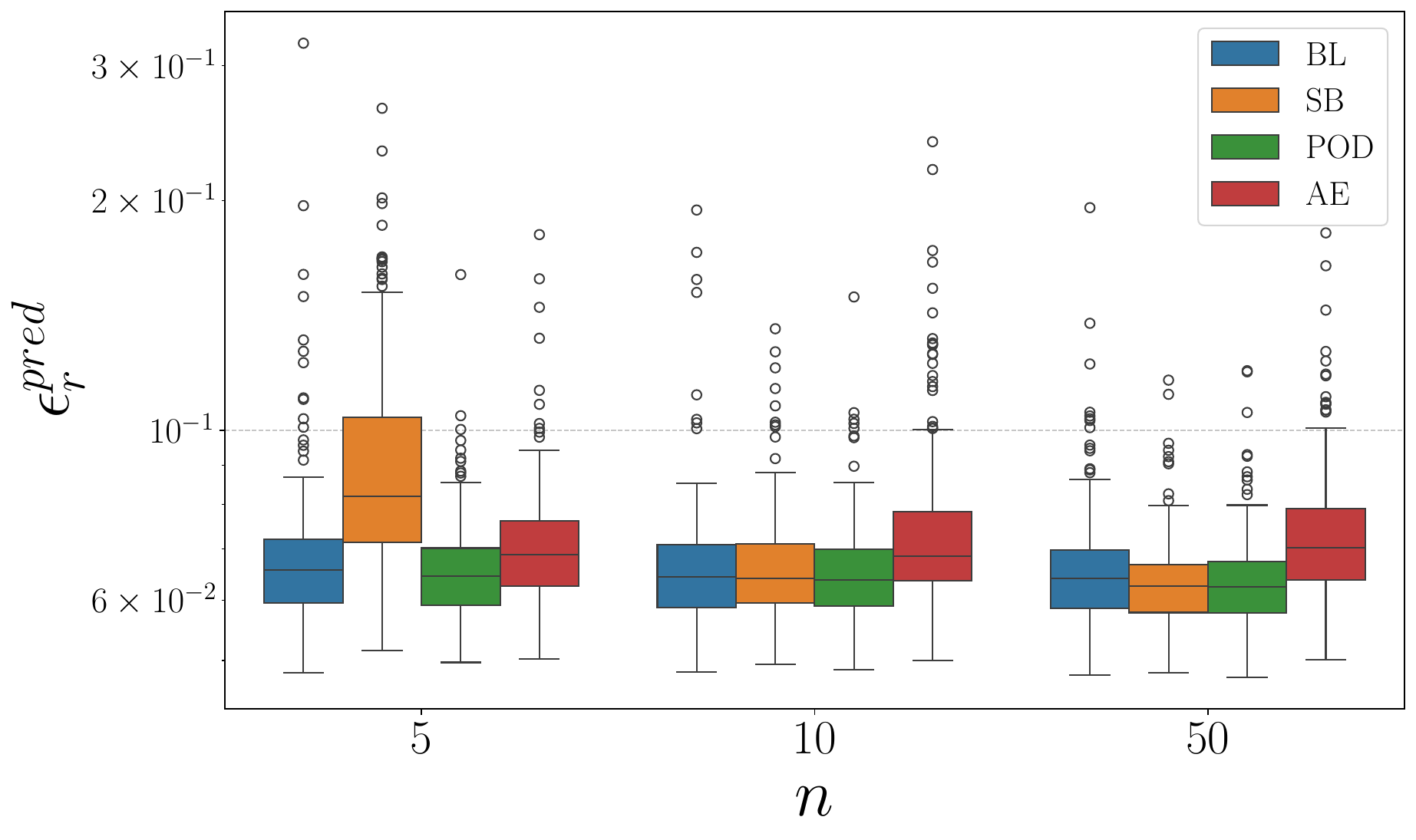}
        \caption{$\mu = 0.05$}
        \label{fig:pred_error_models_N1000R5}
    \end{subfigure}
    \caption{\textbf{Predictive capacity for large dataset size.} Boxplots of the predictive error of different models for $D=1000$. BL: Baseline, SB: Spectral Basis, POD: Proper Orthogonal Decomposition, AE: Autoencoder.}
    \label{fig:pred_error_N1000}
\end{figure}

The results of the predictive and explanatory capacity of the models for the different variables combinations described above are shown in Tables \ref{tab:baseline}, \ref{tab:FFT}, \ref{tab:POD} and \ref{tab:autoencoder}. Each table correspond to one approach, being \ref{tab:baseline} the results of the state of the art \cite{munoz2025application}. The results are stratified by $D$, $\mu$ and $n$. In terms of the predictive capacity, the three quartiles of the prediction error with the validation data are reported, evaluated with Eq. (\ref{eq:predictive_error}), as well as the explanatory error, evaluated using Eq. (\ref{eq:explanatory_error}). Figs. \ref{fig:pred_error_N100} and \ref{fig:pred_error_N1000} summarize the results from the tables for low-noise data ($\mu \in \{0.00, 0.01\}$). 

In terms of the predictive capacity, we observe that when using \emph{a priori} model order reduction, a high number of elements of the basis is required to capture data variability, but the spectral approximation contributes to noise filtering, specially at the small data regime. On the other hand, for \emph{a posteriori} model order reduction, the use of POD-based decoders is competitive with the state of the art for small data sets, whereas for bigger datasets, the approach using the decoder of a previously trained autoencoder is also competitive, also when data is noisy.

In terms of the explanatory capacity, all the presented approaches are competitive with the state of the art, and even improve the discovery of the state equation provided that the dimension of the latent space is high enough.

\begin{table}[h]
\renewcommand{\arraystretch}{1.3} 
\centering
\begin{tabular}{ccccccccc}
\hline
 &  &  &  & $\bs{\epsilon}_r^{\text{pred}}$ &  &  \\
\cline{4-6}
$\bs{D}$ & $\bs{\mu}$ (\%) & $\bs{n}$ &   $\bs{(Q_1)}$ & $\bs{(Q_2)}$ & $\bs{(Q_3)}$ & $\bs{\epsilon}_r^{\text{exp}}$ \\
\hline
\multirow[c]{9}{*}{10} & \multirow[c]{3}{*}{0} & 5 & 8.13$\times 10^{-2}$ & 9.08$\times 10^{-2}$ & 1.00$\times 10^{-1}$ & 1.98$\times 10^{-2}$ \\
 &  & 10 & 7.57$\times 10^{-2}$ & 8.60$\times 10^{-2}$ & 9.62$\times 10^{-2}$ & 2.68$\times 10^{-2}$ \\
 &  & 50 & 7.51$\times 10^{-2}$ & 8.53$\times 10^{-2}$ & 9.55$\times 10^{-2}$ & 2.67$\times 10^{-2}$ \\
\cline{2-7}
 & \multirow[c]{3}{*}{1} & 5 & 1.48$\times 10^{-1}$ & 1.79$\times 10^{-1}$ & 2.10$\times 10^{-1}$ & 5.22$\times 10^{-1}$ \\
 &  & 10 & 7.66$\times 10^{-2}$ & 8.84$\times 10^{-2}$ & 1.00$\times 10^{-1}$ & 5.68$\times 10^{-1}$ \\
 &  & 50 & 7.72$\times 10^{-2}$ & 8.43$\times 10^{-2}$ & 9.14$\times 10^{-2}$ & 4.81$\times 10^{-1}$ \\
\cline{2-7}
 & \multirow[c]{3}{*}{5} & 5 & 1.03$\times 10^{-1}$ & 1.11$\times 10^{-1}$ & 1.19$\times 10^{-1}$ & 9.98$\times 10^{-1}$ \\
 &  & 10 & 1.16$\times 10^{-1}$ & 1.19$\times 10^{-1}$ & 1.22$\times 10^{-1}$ & 9.96$\times 10^{-1}$ \\
 &  & 50 & 1.04$\times 10^{-1}$ & 1.09$\times 10^{-1}$ & 1.14$\times 10^{-1}$ & 9.96$\times 10^{-1}$ \\
\cline{1-7} \cline{2-7}
\multirow[c]{9}{*}{100} & \multirow[c]{3}{*}{0} & 5 & 1.18$\times 10^{-3}$ & 2.99$\times 10^{-3}$ & 9.95$\times 10^{-3}$ & 4.00$\times 10^{-2}$ \\
 &  & 10 & 2.24$\times 10^{-3}$ & 4.29$\times 10^{-3}$ & 9.88$\times 10^{-3}$ & 9.64$\times 10^{-2}$ \\
 &  & 50 & 2.08$\times 10^{-3}$ & 6.78$\times 10^{-3}$ & 1.62$\times 10^{-2}$ & 4.08$\times 10^{-2}$ \\
\cline{2-7}
 & \multirow[c]{3}{*}{1} & 5 & 1.65$\times 10^{-2}$ & 2.36$\times 10^{-2}$ & 5.28$\times 10^{-2}$ & 3.33$\times 10^{-1}$ \\
 &  & 10 & 1.79$\times 10^{-2}$ & 2.46$\times 10^{-2}$ & 3.55$\times 10^{-2}$ & 3.25$\times 10^{-1}$ \\
 &  & 50 & 1.71$\times 10^{-2}$ & 2.13$\times 10^{-2}$ & 3.30$\times 10^{-2}$ & 3.57$\times 10^{-1}$ \\
\cline{2-7}
 & \multirow[c]{3}{*}{5} & 5 & 9.54$\times 10^{-2}$ & 1.18$\times 10^{-1}$ & 1.91$\times 10^{-1}$ & 9.72$\times 10^{-1}$ \\
 &  & 10 & 1.51$\times 10^{-1}$ & 1.82$\times 10^{-1}$ & 2.08$\times 10^{-1}$ & 9.72$\times 10^{-1}$ \\
 &  & 50 & 1.09$\times 10^{-1}$ & 1.37$\times 10^{-1}$ & 2.11$\times 10^{-1}$ & 1.04$\times 10^{0}$ \\
\cline{1-7} \cline{2-7}
\multirow[c]{9}{*}{1000} & \multirow[c]{3}{*}{0} & 5 & 1.13$\times 10^{-4}$ & 1.60$\times 10^{-4}$ & 2.87$\times 10^{-4}$ & 3.55$\times 10^{-2}$ \\
 &  & 10 & 1.93$\times 10^{-4}$ & 2.33$\times 10^{-4}$ & 3.35$\times 10^{-4}$ & 2.88$\times 10^{-2}$ \\
 &  & 50 & 1.44$\times 10^{-4}$ & 1.82$\times 10^{-4}$ & 2.88$\times 10^{-4}$ & 2.39$\times 10^{-2}$ \\
\cline{2-7}
 & \multirow[c]{3}{*}{1} & 5 & 1.17$\times 10^{-2}$ & 1.26$\times 10^{-2}$ & 1.35$\times 10^{-2}$ & 1.19$\times 10^{-1}$ \\
 &  & 10 & 1.17$\times 10^{-2}$ & 1.26$\times 10^{-2}$ & 1.35$\times 10^{-2}$ & 4.26$\times 10^{-2}$ \\
 &  & 50 & 1.17$\times 10^{-2}$ & 1.26$\times 10^{-2}$ & 1.35$\times 10^{-2}$ & 3.08$\times 10^{-2}$ \\
\cline{2-7}
 & \multirow[c]{3}{*}{5} & 5 & 5.95$\times 10^{-2}$ & 6.57$\times 10^{-2}$ & 7.21$\times 10^{-2}$ & 8.75$\times 10^{-1}$ \\
 &  & 10 & 5.87$\times 10^{-2}$ & 6.43$\times 10^{-2}$ & 7.08$\times 10^{-2}$ & 8.43$\times 10^{-1}$ \\
 &  & 50 & 5.85$\times 10^{-2}$ & 6.41$\times 10^{-2}$ & 6.98$\times 10^{-2}$ & 9.27$\times 10^{-1}$ \\
\cline{1-7} \cline{3-7}
\end{tabular}
\vspace{0.5em}
\caption{\textbf{Baseline approach performance.} Predictive and explanatory capacity for different values of data size, noise and latent space dimension.}
\label{tab:baseline}
\end{table}

\begin{table}[h]
\renewcommand{\arraystretch}{1.3} 
\centering
\begin{tabular}{ccccccccc}
\hline
 &  &  &  & $\bs{\epsilon}_r^{\text{pred}}$ &  &  \\
\cline{4-6}
$\bs{D}$ & $\bs{\mu}$ (\%) & $\bs{n}$ &   $\bs{(Q_1)}$ & $\bs{(Q_2)}$ & $\bs{(Q_3)}$ & $\bs{\epsilon}_r^{\text{exp}}$ \\
\hline
\multirow[c]{9}{*}{10} & \multirow[c]{3}{*}{0} & 5 & 5.27$\times 10^{-2}$ & 5.83$\times 10^{-2}$ & 6.40$\times 10^{-2}$ & 2.04$\times 10^{0}$ \\
 &  & 10 & 3.63$\times 10^{-2}$ & 4.09$\times 10^{-2}$ & 4.56$\times 10^{-2}$ & 7.59$\times 10^{-1}$ \\
 &  & 50 & 6.57$\times 10^{-2}$ & 6.93$\times 10^{-2}$ & 7.29$\times 10^{-2}$ & 1.15$\times 10^{-1}$ \\
\cline{2-7}
 & \multirow[c]{3}{*}{1} & 5 & 5.25$\times 10^{-2}$ & 5.85$\times 10^{-2}$ & 6.45$\times 10^{-2}$ & 8.53$\times 10^{-1}$ \\
 &  & 10 & 3.85$\times 10^{-2}$ & 4.76$\times 10^{-2}$ & 5.67$\times 10^{-2}$ & 3.32$\times 10^{-1}$ \\
 &  & 50 & 4.88$\times 10^{-2}$ & 4.92$\times 10^{-2}$ & 4.95$\times 10^{-2}$ & 1.77$\times 10^{-1}$ \\
\cline{2-7}
 & \multirow[c]{3}{*}{5} & 5 & 7.61$\times 10^{-2}$ & 8.14$\times 10^{-2}$ & 8.67$\times 10^{-2}$ & 1.11$\times 10^{0}$ \\
 &  & 10 & 7.95$\times 10^{-2}$ & 8.72$\times 10^{-2}$ & 9.48$\times 10^{-2}$ & 7.84$\times 10^{-1}$ \\
 &  & 50 & 1.17$\times 10^{-1}$ & 1.45$\times 10^{-1}$ & 1.74$\times 10^{-1}$ & 9.57$\times 10^{-1}$ \\
\cline{1-7} \cline{2-7}
\multirow[c]{9}{*}{100} & \multirow[c]{3}{*}{0} & 5 & 4.56$\times 10^{-2}$ & 6.43$\times 10^{-2}$ & 9.38$\times 10^{-2}$ & 1.03$\times 10^{0}$ \\
 &  & 10 & 1.19$\times 10^{-2}$ & 2.04$\times 10^{-2}$ & 2.75$\times 10^{-2}$ & 3.92$\times 10^{-1}$ \\
 &  & 50 & 1.88$\times 10^{-3}$ & 2.72$\times 10^{-3}$ & 9.91$\times 10^{-3}$ & 8.60$\times 10^{-2}$ \\
\cline{2-7}
 & \multirow[c]{3}{*}{1} & 5 & 4.83$\times 10^{-2}$ & 6.42$\times 10^{-2}$ & 9.47$\times 10^{-2}$ & 8.15$\times 10^{-1}$ \\
 &  & 10 & 1.86$\times 10^{-2}$ & 2.50$\times 10^{-2}$ & 3.12$\times 10^{-2}$ & 4.56$\times 10^{-1}$ \\
 &  & 50 & 1.31$\times 10^{-2}$ & 1.58$\times 10^{-2}$ & 2.04$\times 10^{-2}$ & 5.09$\times 10^{-2}$ \\
\cline{2-7}
 & \multirow[c]{3}{*}{5} & 5 & 7.89$\times 10^{-2}$ & 9.04$\times 10^{-2}$ & 1.17$\times 10^{-1}$ & 6.85$\times 10^{-1}$ \\
 &  & 10 & 6.54$\times 10^{-2}$ & 7.11$\times 10^{-2}$ & 7.63$\times 10^{-2}$ & 6.35$\times 10^{-1}$ \\
 &  & 50 & 6.48$\times 10^{-2}$ & 6.76$\times 10^{-2}$ & 7.39$\times 10^{-2}$ & 4.69$\times 10^{-1}$ \\
\cline{1-7} \cline{2-7}
\multirow[c]{9}{*}{1000} & \multirow[c]{3}{*}{0} & 5 & 4.12$\times 10^{-2}$ & 5.28$\times 10^{-2}$ & 7.77$\times 10^{-2}$ & 7.05$\times 10^{-1}$ \\
 &  & 10 & 6.53$\times 10^{-3}$ & 1.22$\times 10^{-2}$ & 1.94$\times 10^{-2}$ & 5.47$\times 10^{-1}$ \\
 &  & 50 & 3.10$\times 10^{-4}$ & 4.75$\times 10^{-4}$ & 7.33$\times 10^{-4}$ & 4.65$\times 10^{-2}$ \\
\cline{2-7}
 & \multirow[c]{3}{*}{1} & 5 & 4.26$\times 10^{-2}$ & 5.37$\times 10^{-2}$ & 7.92$\times 10^{-2}$ & 9.82$\times 10^{-1}$ \\
 &  & 10 & 1.40$\times 10^{-2}$ & 1.77$\times 10^{-2}$ & 2.37$\times 10^{-2}$ & 1.12$\times 10^{0}$ \\
 &  & 50 & 1.16$\times 10^{-2}$ & 1.24$\times 10^{-2}$ & 1.34$\times 10^{-2}$ & 3.37$\times 10^{-2}$ \\
\cline{2-7}
 & \multirow[c]{3}{*}{5} & 5 & 7.14$\times 10^{-2}$ & 8.20$\times 10^{-2}$ & 1.04$\times 10^{-1}$ & 8.98$\times 10^{-1}$ \\
 &  & 10 & 5.95$\times 10^{-2}$ & 6.40$\times 10^{-2}$ & 7.11$\times 10^{-2}$ & 4.72$\times 10^{-1}$ \\
 &  & 50 & 5.78$\times 10^{-2}$ & 6.25$\times 10^{-2}$ & 6.67$\times 10^{-2}$ & 5.22$\times 10^{-1}$ \\
\cline{1-7} \cline{3-7}
\end{tabular}
\vspace{0.5em}
\caption{\textbf{Fourier approach performance.} Predictive and explanatory capacity for different values of data size, noise and latent space dimension.}
\label{tab:FFT}
\end{table}

\begin{table}[h]
\renewcommand{\arraystretch}{1.3} 
\centering
\begin{tabular}{ccccccccc}
\hline
 &  &  &  & $\bs{\epsilon}_r^{\text{pred}}$ &  &  \\
\cline{4-6}
$\bs{D}$ & $\bs{\mu}$ (\%) & $\bs{n}$ &   $\bs{(Q_1)}$ & $\bs{(Q_2)}$ & $\bs{(Q_3)}$ & $\bs{\epsilon}_r^{\text{exp}}$ \\
\hline
\multirow[c]{9}{*}{10} & \multirow[c]{3}{*}{0} & 5 & 3.32$\times 10^{-2}$ & 4.32$\times 10^{-2}$ & 5.32$\times 10^{-2}$ & 1.91$\times 10^{-2}$ \\
\cline{2-7}
 & \multirow[c]{3}{*}{1} & 5 & 4.50$\times 10^{-2}$ & 5.16$\times 10^{-2}$ & 5.81$\times 10^{-2}$ & 5.17$\times 10^{-1}$ \\
\cline{2-7}
 & \multirow[c]{3}{*}{5} & 5 & 8.59$\times 10^{-2}$ & 9.33$\times 10^{-2}$ & 1.01$\times 10^{-1}$ & 8.75$\times 10^{-1}$ \\
\cline{1-7} \cline{2-7}
\multirow[c]{9}{*}{100} & \multirow[c]{3}{*}{0} & 5 & 2.49$\times 10^{-3}$ & 3.69$\times 10^{-3}$ & 1.03$\times 10^{-2}$ & 4.06$\times 10^{-2}$ \\
 &  & 10 & 1.25$\times 10^{-3}$ & 3.11$\times 10^{-3}$ & 9.27$\times 10^{-3}$ & 9.41$\times 10^{-2}$ \\
 &  & 50 & 2.09$\times 10^{-3}$ & 4.66$\times 10^{-3}$ & 1.06$\times 10^{-2}$ & 4.45$\times 10^{-2}$ \\
\cline{2-7}
 & \multirow[c]{3}{*}{1} & 5 & 1.40$\times 10^{-2}$ & 1.76$\times 10^{-2}$ & 2.26$\times 10^{-2}$ & 2.03$\times 10^{-1}$ \\
 &  & 10 & 1.42$\times 10^{-2}$ & 1.75$\times 10^{-2}$ & 2.42$\times 10^{-2}$ & 3.84$\times 10^{-1}$ \\
 &  & 50 & 1.37$\times 10^{-2}$ & 1.69$\times 10^{-2}$ & 2.36$\times 10^{-2}$ & 2.18$\times 10^{-1}$ \\
\cline{2-7}
 & \multirow[c]{3}{*}{5} & 5 & 6.38$\times 10^{-2}$ & 6.83$\times 10^{-2}$ & 7.23$\times 10^{-2}$ & 5.98$\times 10^{-1}$ \\
 &  & 10 & 6.49$\times 10^{-2}$ & 7.72$\times 10^{-2}$ & 8.45$\times 10^{-2}$ & 9.63$\times 10^{-1}$ \\
 &  & 50 & 6.40$\times 10^{-2}$ & 7.53$\times 10^{-2}$ & 8.10$\times 10^{-2}$ & 9.69$\times 10^{-1}$ \\
\cline{1-7} \cline{2-7}
\multirow[c]{9}{*}{1000} & \multirow[c]{3}{*}{0} & 5 & 8.34$\times 10^{-4}$ & 1.22$\times 10^{-3}$ & 1.98$\times 10^{-3}$ & 4.36$\times 10^{-2}$ \\
 &  & 10 & 1.86$\times 10^{-4}$ & 2.46$\times 10^{-4}$ & 3.86$\times 10^{-4}$ & 2.70$\times 10^{-2}$ \\
 &  & 50 & 1.54$\times 10^{-4}$ & 2.04$\times 10^{-4}$ & 3.24$\times 10^{-4}$ & 2.38$\times 10^{-2}$ \\
\cline{2-7}
 & \multirow[c]{3}{*}{1} & 5 & 1.16$\times 10^{-2}$ & 1.26$\times 10^{-2}$ & 1.36$\times 10^{-2}$ & 1.02$\times 10^{-1}$ \\
 &  & 10 & 1.16$\times 10^{-2}$ & 1.24$\times 10^{-2}$ & 1.33$\times 10^{-2}$ & 1.40$\times 10^{-1}$ \\
 &  & 50 & 1.15$\times 10^{-2}$ & 1.23$\times 10^{-2}$ & 1.35$\times 10^{-2}$ & 9.11$\times 10^{-2}$ \\
\cline{2-7}
 & \multirow[c]{3}{*}{5} & 5 & 5.90$\times 10^{-2}$ & 6.45$\times 10^{-2}$ & 7.02$\times 10^{-2}$ & 6.05$\times 10^{-1}$ \\
 &  & 10 & 5.90$\times 10^{-2}$ & 6.37$\times 10^{-2}$ & 6.99$\times 10^{-2}$ & 8.43$\times 10^{-1}$ \\
 &  & 50 & 5.77$\times 10^{-2}$ & 6.24$\times 10^{-2}$ & 6.73$\times 10^{-2}$ & 6.56$\times 10^{-1}$ \\
\cline{1-7} \cline{3-7}
\end{tabular}
\vspace{0.5em}
\caption{\textbf{POD approach performance.} Predictive and explanatory capacity for different values of data size, noise and latent space dimension.}
\label{tab:POD}
\end{table}

\begin{table}[h]
\renewcommand{\arraystretch}{1.3} 
\centering
\begin{tabular}{ccccccccc}
\hline
 &  &  &  & $\bs{\epsilon}_r^{\text{pred}}$ &  &  \\
\cline{4-6}
$\bs{D}$ & $\bs{\mu}$ (\%) & $\bs{n}$ &   $\bs{(Q_1)}$ & $\bs{(Q_2)}$ & $\bs{(Q_3)}$ & $\bs{\epsilon}_r^{\text{exp}}$ \\
\hline
\multirow[c]{9}{*}{10} & \multirow[c]{3}{*}{0} & 5 & 7.37$\times 10^{-2}$ & 1.00$\times 10^{-1}$ & 1.27$\times 10^{-1}$ & 5.09$\times 10^{0}$ \\
 &  & 10 & 9.13$\times 10^{-2}$ & 1.04$\times 10^{-1}$ & 1.17$\times 10^{-1}$ & 9.91$\times 10^{-1}$ \\
 &  & 50 & 1.00$\times 10^{-1}$ & 1.19$\times 10^{-1}$ & 1.37$\times 10^{-1}$ & 1.11$\times 10^{0}$ \\
\cline{2-7}
 & \multirow[c]{3}{*}{1} & 5 & 8.89$\times 10^{-2}$ & 1.12$\times 10^{-1}$ & 1.34$\times 10^{-1}$ & 1.24$\times 10^{0}$ \\
 &  & 10 & 8.83$\times 10^{-2}$ & 1.00$\times 10^{-1}$ & 1.12$\times 10^{-1}$ & 4.97$\times 10^{-1}$ \\
 &  & 50 & 1.07$\times 10^{-1}$ & 1.23$\times 10^{-1}$ & 1.40$\times 10^{-1}$ & 1.14$\times 10^{0}$ \\
\cline{2-7}
 & \multirow[c]{3}{*}{5} & 5 & 1.10$\times 10^{-1}$ & 1.19$\times 10^{-1}$ & 1.28$\times 10^{-1}$ & 1.40$\times 10^{0}$ \\
 &  & 10 & 1.08$\times 10^{-1}$ & 1.14$\times 10^{-1}$ & 1.21$\times 10^{-1}$ & 1.81$\times 10^{0}$ \\
 &  & 50 & 1.05$\times 10^{-1}$ & 1.15$\times 10^{-1}$ & 1.24$\times 10^{-1}$ & 7.11$\times 10^{-1}$ \\
\cline{1-7} \cline{2-7}
\multirow[c]{9}{*}{100} & \multirow[c]{3}{*}{0} & 5 & 4.91$\times 10^{-3}$ & 8.03$\times 10^{-3}$ & 7.18$\times 10^{-2}$ & 2.18$\times 10^{-1}$ \\
 &  & 10 & 6.96$\times 10^{-3}$ & 1.59$\times 10^{-2}$ & 4.42$\times 10^{-2}$ & 6.00$\times 10^{-2}$ \\
 &  & 50 & 7.45$\times 10^{-3}$ & 1.28$\times 10^{-2}$ & 3.17$\times 10^{-2}$ & 9.11$\times 10^{-2}$ \\
\cline{2-7}
 & \multirow[c]{3}{*}{1} & 5 & 1.85$\times 10^{-2}$ & 2.16$\times 10^{-2}$ & 9.95$\times 10^{-2}$ & 7.42$\times 10^{-1}$ \\
 &  & 10 & 1.67$\times 10^{-2}$ & 2.56$\times 10^{-2}$ & 5.70$\times 10^{-2}$ & 1.03$\times 10^{0}$ \\
 &  & 50 & 2.40$\times 10^{-2}$ & 4.47$\times 10^{-2}$ & 7.32$\times 10^{-2}$ & 1.21$\times 10^{0}$ \\
\cline{2-7}
 & \multirow[c]{3}{*}{5} & 5 & 9.88$\times 10^{-2}$ & 1.25$\times 10^{-1}$ & 1.72$\times 10^{-1}$ & 7.36$\times 10^{-1}$ \\
 &  & 10 & 1.07$\times 10^{-1}$ & 1.59$\times 10^{-1}$ & 2.04$\times 10^{-1}$ & 5.97$\times 10^{-1}$ \\
 &  & 50 & 8.40$\times 10^{-2}$ & 1.01$\times 10^{-1}$ & 1.69$\times 10^{-1}$ & 7.07$\times 10^{-1}$ \\
\cline{1-7} \cline{2-7}
\multirow[c]{9}{*}{1000} & \multirow[c]{3}{*}{0} & 5 & 5.09$\times 10^{-4}$ & 6.41$\times 10^{-4}$ & 1.21$\times 10^{-3}$ & 1.82$\times 10^{-2}$ \\
 &  & 10 & 3.23$\times 10^{-4}$ & 5.35$\times 10^{-4}$ & 1.01$\times 10^{-3}$ & 1.92$\times 10^{-2}$ \\
 &  & 50 & 3.75$\times 10^{-4}$ & 5.29$\times 10^{-4}$ & 1.00$\times 10^{-3}$ & 4.09$\times 10^{-2}$ \\
\cline{2-7}
 & \multirow[c]{3}{*}{1} & 5 & 1.19$\times 10^{-2}$ & 1.27$\times 10^{-2}$ & 1.39$\times 10^{-2}$ & 8.79$\times 10^{-2}$ \\
 &  & 10 & 1.19$\times 10^{-2}$ & 1.28$\times 10^{-2}$ & 1.37$\times 10^{-2}$ & 9.97$\times 10^{-2}$ \\
 &  & 50 & 1.18$\times 10^{-2}$ & 1.28$\times 10^{-2}$ & 1.40$\times 10^{-2}$ & 1.25$\times 10^{-1}$ \\
\cline{2-7}
 & \multirow[c]{3}{*}{5} & 5 & 6.26$\times 10^{-2}$ & 6.87$\times 10^{-2}$ & 7.61$\times 10^{-2}$ & 8.52$\times 10^{-1}$ \\
 &  & 10 & 6.36$\times 10^{-2}$ & 6.85$\times 10^{-2}$ & 7.83$\times 10^{-2}$ & 9.77$\times 10^{-1}$ \\
 &  & 50 & 6.38$\times 10^{-2}$ & 7.02$\times 10^{-2}$ & 7.90$\times 10^{-2}$ & 7.92$\times 10^{-1}$ \\
\cline{1-7} \cline{3-7}
\end{tabular}
\vspace{0.5em}
\caption{\textbf{Autoencoder approach performance.} Predictive and explanatory capacity for different values of data size, noise and latent space dimension.}
\label{tab:autoencoder}
\end{table}

\clearpage
\subsubsection{Computational savings}
The main advantage of the proposed approach lies on its scalability and on its reduction in computational cost. By using embedding-oriented PGNNIVs, our method achieves faster execution times respecting to the original method, independently of the dimension of the output variable. These computational savings and  quantitative comparisons to the baseline method are reported in Table \ref{tab:mean_std_time}, where we show the speed-up increase with respect to the baseline model, reporting the mean and standard deviation of the acceleration rates across the different $D$, grouped by the latent space size $n$ and the noise $\mu$. Values lower than $1.0$ indicate a speed-up. Across various configurations, the table demonstrates the speed-ups of the different methods. It is important to note that this acceleration is achieved for $m=\sqrt{N} = 10$, but, as demonstrated in section \ref{subsec::model_complexity}, the model complexity is $\mathcal{O}(m^2)$ so this acceleration would become critical as the dimension of the output $N$ increases.


\begin{table}[h!]
\renewcommand{\arraystretch}{1.3} 
\centering
\begin{tabular}{lll}
\hline
$\bs{D}$ & \textbf{Model} & \textbf{Acceleration rate} \\
\hline
     & Fourier      & $0.94 \pm 0.03 \ ^{***}$ \\
 10  & POD          & $0.92 \pm 0.01 \ ^{*}$ \\
     & Autoencoder  & $1.05 \pm 0.03$ \\
\hline
     & Fourier      & $0.95 \pm 0.04 \ ^{***}$ \\
100  & POD          & $0.94 \pm 0.03 \ ^{***}$ \\
     & Autoencoder  & $1.10 \pm 0.04$ \\
\hline
     & Fourier      & $0.94 \pm 0.02 \ ^{***}$ \\
1000 & POD          & $0.94 \pm 0.01 \ ^{***}$ \\
     & Autoencoder  & $0.61 \pm 0.00 \ ^{***}$ \\
\hline
\end{tabular}
\vspace{0.5em}
\caption{\textbf{Speed-up analysis.} Mean and standard deviation of the time reduction factor. Statistical significance was assessed using the Mann–Whitney U test \cite{hettmansperger2010robust}. Asterisks indicate the significance levels in terms of $p$-values: $(^*)$ for $p < 0.05$, $(^{**})$ for $p < 0.01$, and $(^{***})$ for $p < 0.001$.}
\label{tab:mean_std_time}
\end{table}

\subsubsection{Overfitting evaluation}
\begin{figure}[h!]
    \centering
    
    \begin{subfigure}[b]{0.47\textwidth}
        \centering
        \includegraphics[width=\textwidth]{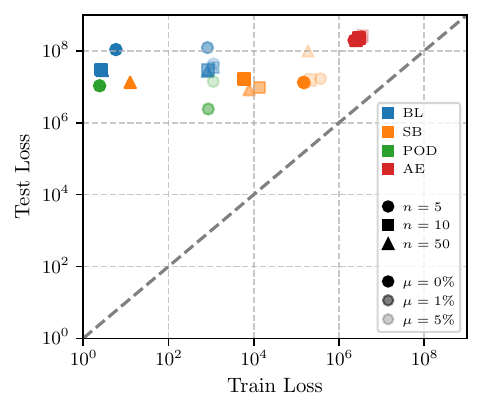}
        \caption{$D = 10$}
        \label{fig:overfitting_N10}
    \end{subfigure}
    \quad \quad
    \begin{subfigure}[b]{0.47\textwidth}
        \centering
        \includegraphics[width=\textwidth]{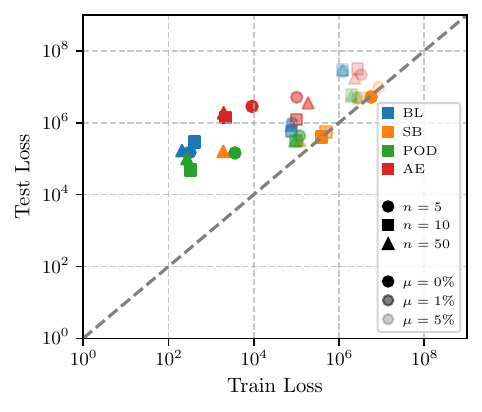}
        \caption{$D = 100$}
        \label{fig:overfitting_N100}
    \end{subfigure}\\
    \vspace{0.5cm}
    \begin{subfigure}[b]{0.47\textwidth}
        \centering
        \includegraphics[width=\textwidth]{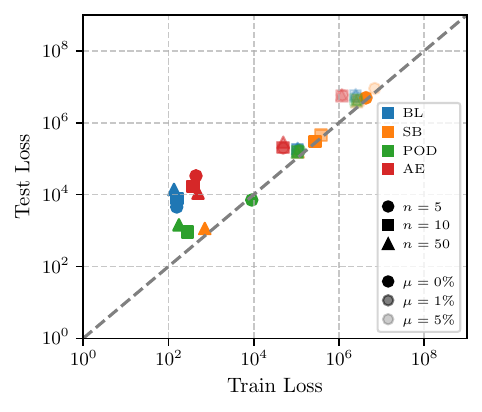}
        \caption{$D = 1000$}
        \label{fig:overfitting_N1000}
    \end{subfigure}
    \caption{\textbf{Overfitting evaluation.} Scatterplot of the train vs test loss for different dataset sizes. The closer the approaches are to the diagonal, the less overfitting. BL: Baseline, SB: Spectral Basis, POD: Proper Orthogonal Decomposition, AE: Autoencoder.}
    \label{fig:overfitting_plots}
\end{figure}

In order to understand how the models deal with overfitting, it is essential to analyze their performance on both training and validation datasets. By comparing the train data loss and the test data loss across the datasets, it is possible to explore whether a model is overfitted or not. In Fig. \ref{fig:overfitting_plots} three plots showing train and test loss to find out how the different models behave in terms of overfitting. Notably, regardless of overall performance, the three proposed approaches are in general less prone to overfitting.

\subsection{Knowledge transfer} \label{subsecc::results_transfer_learning}

To leverage knowledge acquired previously, we use one of the baseline models trained in the previous section. Although the baseline model is used in this instance, an autoencoder-based approach could alternatively be adopted. The essential requirement is that the chosen method must support the joint training of both the encoder and the decoder components. The model chosen is the one trained for $D = 1000$ samples, a latent space of $n=10$ and zero noise ($\mu = 0$) and whose relationship between $K$ and $u$ follows a quadratic dependence (Material 1). We consider two scenarios in our transfer learning approach:
\begin{enumerate}
    \item \textbf{Retraining a new model with a pre-trained encoder (Encoder Frozen):}  
    In this scenario, we keep the encoder component of the pre-trained model, which has already learned a meaningful representation of the input data. The encoder's weights are frozen, so they are not updated during the training. The decoder and the explanatory network are now retrained, but not from random reinitialized weights. Instead, they have been retrained from the previously obtained tentative solution, rather than being trained entirely from scratch.

    \item \textbf{Fine-tuning the pretrained encoder (Encoder Trainable):}  
    In the second scenario, we initialize the model using the pre-trained encoder. However, unlike the first case, we allow the encoder weights to be fine-tuned during training on the new task.  The decoder and the explanatory network are trained, but with the advantage that the encoder has already learned a meaningful representation of the input data. As a result, the training process starts closer to an optimal solution compared to training the entire model from scratch. Moreover, since the encoder is also fine-tuned, the model allows an improved optimization, relative to a standard transfer learning approach."
\end{enumerate}

\begin{figure}[h!]
    \centering
    \begin{subfigure}[t]{0.47\linewidth}
        \centering
        \includegraphics[width=\linewidth]{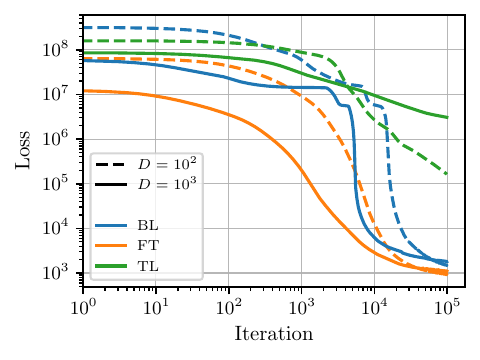}
        \caption{Loss evolution during training.}
        \label{fig:transfer_learning_convergence}
    \end{subfigure}
    \quad \quad
    \begin{subfigure}[t]{0.47\linewidth}
        \centering
        \includegraphics[width=\linewidth]{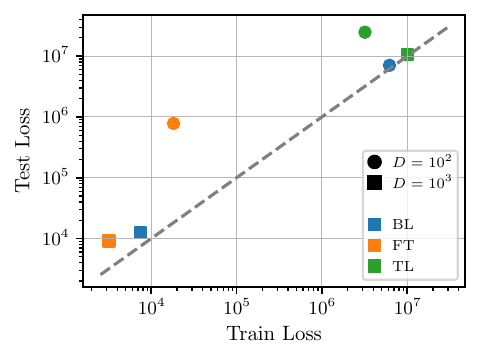} 
        \caption{Scatter plot of train vs test loss.}
        \label{fig:transfer_learning_overfitting}
    \end{subfigure}
    \caption{\textbf{Analysis of the training performance.} Comparison of the train and test loss for the different transfer learning approaches and different dataset sizes. BL: Baseline, FT: Fine-tuning, TL: Transfer Learning.}
    \label{fig:transfer_learning_cumtime}
\end{figure}

We use two new datasets of $D'=100$ and $D'=1000$ samples, each containing a material whose relationship between $K$ and $u$ follows a sigmoid dependence (Material 2). A comparison is made with a model trained from scratch with the same dataset (baseline). In order to compare the three different approaches to obtain the constitutive behavior of Material 2, each model is trained for $10^5$ epochs, with a learning rate $\text{lr} = 3 \times 10^{-4}$ using Adam method. For $D' = 100$, the baseline training time is $18.39$ minutes. Fine-tuning achieves a slight speedup of $1.9\%$, reducing the training time to approximately $18.05$ minutes. In contrast, transfer learning results in a more substantial reduction to $15.93$ minutes, corresponding to a $13.4\%$ improvement. For $D' = 1000$, the baseline training time increases significantly to $168.93$ minutes. Fine-tuning provides a modest acceleration of $1.3\%$, decreasing the time to $166.67$ minutes. Transfer learning, however, achieves a much larger gain, reducing the training time to $126.91$ minutes, which represents a $24.9\%$ improvement. Figure~\ref{fig:transfer_learning_cumtime} shows the cumulative training time per epoch for the $D' = 1000$ dataset across the different methods.

\begin{figure}[h!]
    \centering\includegraphics{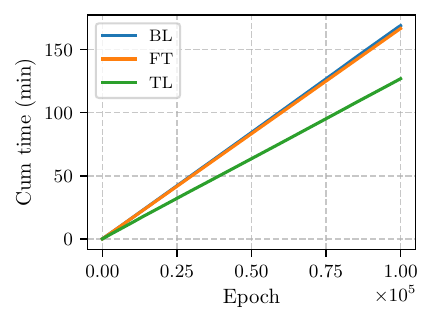}
    \caption{\textbf{Training time evaluation.} Duration of training  throughout the epochs for $D = 1000$. BL: Baseline, FT: Fine-tuning, TL: Transfer Learning.}
    \label{fig:transfer_learning_cumtime_N1000}
\end{figure}

\begin{figure}[h!]
  \centering
  \begin{subfigure}{0.48\textwidth}
    \includegraphics[width=\linewidth]{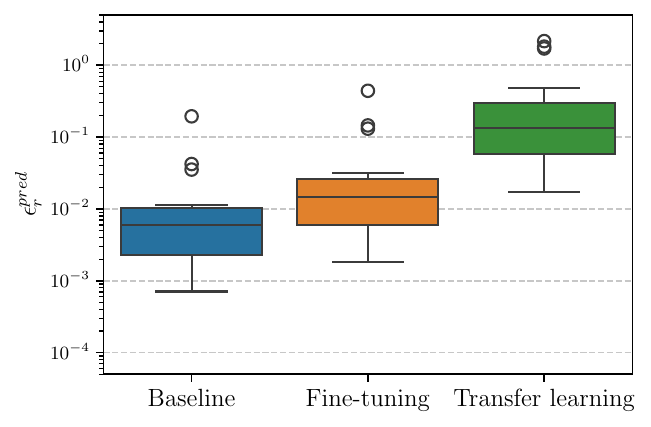}
    \caption{$D=100$.}
    \label{subfig:TL_predictive_error_N100}
  \end{subfigure}
  \quad
  \begin{subfigure}{0.48\textwidth}
    \includegraphics[width=\linewidth]{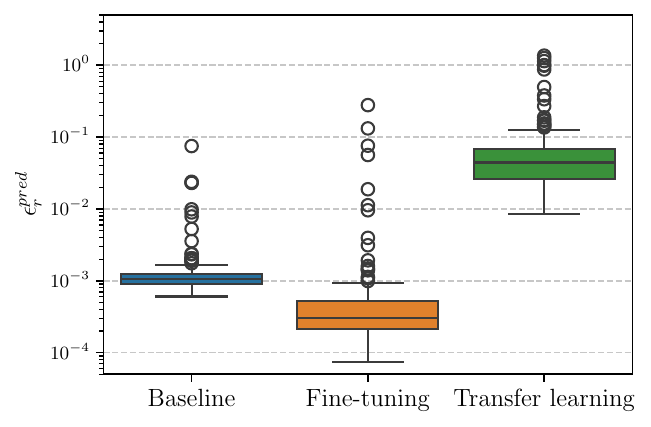}
    \caption{$D=1000$.}
    \label{subfig:transfer_learning_overfitting}
  \end{subfigure}
  \caption{\textbf{Predictive capacity} Boxplot of the predictive errors for the different approaches and different dataset sizes.}
  \label{fig:transfer_learning_new_material_predictive_error}
\end{figure}

\begin{figure}[h!]
  \centering
  \begin{subfigure}{0.46\textwidth}
    \includegraphics[width=\linewidth]{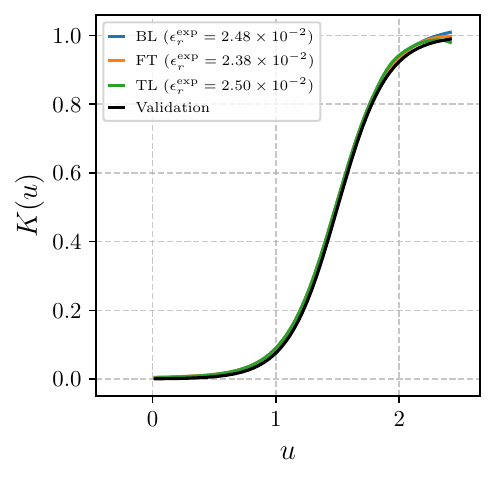}
    \caption{$D=100$.}
    \label{subfig:transfer_learning_N100}
  \end{subfigure}
  \hspace{1cm}
  \begin{subfigure}{0.46\textwidth}
    \includegraphics[width=\linewidth]{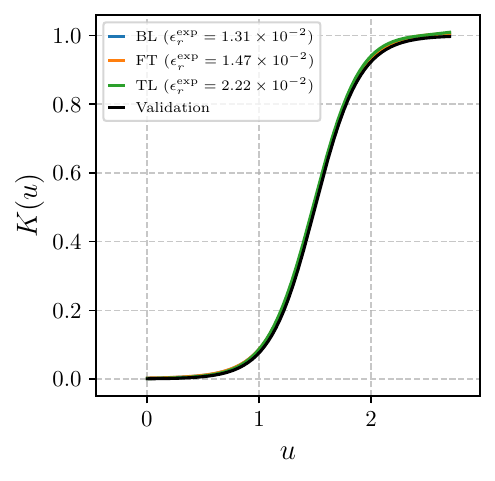}
    \caption{$D=1000$.}
    \label{subfig:transfer_learning_N1000}
  \end{subfigure}
  \caption{\textbf{Explanatory capacity.} New material constitutive behavior unveiling using the different training strategies proposed and different dataset sizes. BL: Baseline, FT: Fine-tuning, TL: Transfer Learning.}
  \label{fig:transfer_learning_new_material_dependence}
\end{figure}

We compare the performance of these two approaches with the third one trained from scratch using the new material. The predictive and explanatory capabilities of all three methods are summarized in Figure~\ref{fig:transfer_learning_new_material_predictive_error}. Furthermore, the learned constitutive relationship $K(u)$ for the new material is visualized in Figure~\ref{fig:transfer_learning_new_material_dependence}, where the explanatory error is also shown. Transfer learning demonstrates to be a compromise between computational cost and accuracy, while fine-tuning achieves better accuracy in predictions when the dataset size is large enough, while keeping a similar accuracy in unveiling the constitutive equation.


\section{Discussion} \label{secc::discussion}
The PGNNIV framework presents a novel approach to combining physical knowledge and machine learning techniques to solve problems in continuum physics, and, as an extra feature, to discover the internal behavior of a system directly from data. This is illustrated here using the stationary diffusion equation, which is simple yet illustrative. However, scalability in terms of higher dimensions and finer discretizations, is one of its main limitations. The presented work illustrates two different strategies for addressing this issue and attempting to solve it: reducing the number of trainable parameters by substituting part of the model with less computationally expensive operations, and a transfer of the learned knowledge from previous experiments into new ones.

\subsection{Parametric analysis} \label{subsecc::parametric_analysis}
\paragraph{Embedding models.} In the case of noise-free data, the relative performance of the models depends on the number of available data. For $D = 10$, the most accurate model in terms of prediction is the POD-PGNNIV, which also has the best results in terms explanatory capability. For $D = 100$, the POD model remains the most accurate in prediction, while the baseline model outperforms the rest in terms of explanatory capability. When increasing to $D = 1000$, the baseline model becomes the most accurate for prediction, whereas the autoencoder achieves the best performance in terms of explanation. However, all these differences are not significant enough to decisively choose one model over another. Across all cases, both the prediction and explanatory errors are of the same order of magnitude, so these metrics do not offer a clear criterion for model selection. While the explanatory capacity of the models remains robust even with limited data, the ability to generalize the solution of the predictive model, still depends of a big enough dataset. In terms of overfitting and generalization, as shown in Fig.(\ref{subfig:transfer_learning_overfitting}), increasing the amount of training data leads to a smaller gap between training and testing losses, as could be anticipated. This indicates that the different approaches benefit from larger datasets to improve their generalization to unseen data. 

Secondly, regarding noise tolerance, the performance of the different models under a $1\%$ of noise degrades but remains within acceptable limits around $1\%$ error. Under this noise level, POD is the most effective at filtering noise, for $D = 100$ and for $D = 1000$. At $5\%$ noise, none of the models achieves good performance. However, the use of Fourier spectral basis achieves the best results in terms of prediction but not explanation—it performs better than the others, but still fails to deliver explanatory capability.

Thirdly, how the training time of each model compares to the baseline model is analyzed in Table \ref{tab:mean_std_time}. For small datasets, the use of the autoencoder is not worthy from a computational point of view.. However, when the dataset size increases to $D = 1000$, it shows a significant reduction in computation time, outperforming the other methods. While the autoencoder does not outperform any of the other models in terms of predictive or explanatory capabilities, its main advantage lies in training speed when large amounts of data are available.

Finally, the number of modes determines how much detail is captured by the model. In the noise-free case its impact varies across methods: for the baseline model, increasing the number of modes does not necessarily lead to better performance (in fact, for $D = 100$ and $1000$, the ordering of performance is 50 modes $>$ 5 modes $>$ 10 modes). Also, for the autoencoder, no consistent pattern is observed with respect to the number of modes. In contrast, for the Fourier and POD models, an increase in the number of modes consistently results in improved performance. This is expected, as a higher number of modes allows the models to capture finer details in the data, leading to more accurate predictions and, therefore, in the explanatory network.

In summary, the three approaches are much more competitive than the state of the art in terms of the model complexity and scalability, while maintaining a good overall performance for state equations discovery and even improving system prediction in certain scenarios (for instance, spectral approaches for small data regimes and uncertainty).

\paragraph{Transfer learning.} Once a PGNNIV has been trained on a dataset, the prediction network can be interpreted as having isolated the underlying information specific to the experimental conditions from which the data were obtained. Consequently, some of the elements of the PGNNIV can be employed for the characterization of a new material. Rather than being trained from scratch, it will be fine-tuned by leveraging the previously acquired knowledge. As shown in Fig. (\ref{fig:transfer_learning_cumtime_N1000}), the total training time for both training from scratch and fine-tuning is approximately the same, as the number of trainable parameters remains unchanged. However, fine-tuning achieves faster convergence. Therefore, in addition to being slightly more time-efficient, it reaches an optimal solution more rapidly than training from scratch, and even better, because the final loss after the whole training is lower. On the other hand, although the convergence of the transfer learning model is not as effective as that of the other two approaches, its training time is drastically shorter.  Recall that none of these approaches suffer from scalability issues of state of the art PGNNIV.

Regarding the performance of the models, in terms of predictive capacity, the transfer learning model performs significantly worse than the other two approaches. This explains its higher loss values, as the solution error is heavily penalized, as mentioned in section \ref{subsec::numerical_experiments}. The reduced performance is due to that the frozen hyperparameters, which are not updated with the new data, belong to the predictive network, so the predictive result will be worse. Consequently, the effectiveness of transfer learning strongly depends on the quality of the original model: the better it is trained, the better the results obtained through transfer learning. Nevertheless, in terms of explanatory capacity, the three models obtain an error of the same order (although the transfer learning model is always the one with a worse performance). Additionally, while increasing the amount of data generally improves model performance, the explanatory capacity of the transfer learning model remains nearly constant. This is because its prediction error does not decrease sufficiently with more data, limiting any significant improvement in its explanatory ability. In short, transfer learning and fine tuning may be envisaged as low-cost alternatives for unravelling purposes adapted to new materials tested under similar experimental setups. 

In terms of overfitting and generalization (see Fig.\ref{subfig:transfer_learning_overfitting}), increasing the amount of training data leads to a smaller gap between training and testing errors. This indicates that the models benefit from larger datasets to improve their generalization to unseen data. While the explanatory capacity of the models remains robust even with limited data, the ability to generalize the solution across unseen data remains dependent of a big enough dataset. In that sense, the choice between the transfer learning and fine-tuning approach may be context dependent and goal oriented. 

\subsection{Limitations and future work} \label{subsecc::limitations_future_work}

Despite its contributions, the presented approach has some limitations that merit further investigation. As is common in the field of ML, one of the main challenges is the substantial volume of data required for model development. We have illustrated the impact of data volume all along the study (c.f. Tables \ref{tab:baseline}, \ref{tab:FFT}, \ref{tab:POD}, \ref{tab:autoencoder} and Figs. \ref{fig:pred_error_N100}, \ref{fig:pred_error_N1000}). 

Besides, one of the key limitations of this methodology is that the loss function contains numerous penalty parameters, that play the role of additional hyperparameters. Consequently, the quality of the solution is highly dependent on them. Some approaches, such as those outlined by \cite{xiang2022self}, propose the use of probabilistic models to define the self-adaptive loss function through adaptive weights for each loss term. 

Furthermore, it should be noted that the methodology has thus far been restricted to simple geometries, namely squares (or in general, hypercubes) with regular meshes. However, this approach can be further enhanced by integrating PGNNIV with the Finite Elements Method (FEM), allowing to apply this methodology to more complex systems. Indeed, a key enabler of this integration is the use of spectral decomposition techniques, as employed in spectral-embedding PGNNIV, that is already foundational to FEM formulation, which provides a natural bridge between PGNNIV and FEM (shape function are local representation basis \cite{boyd2001}). 

Finally, the universal approximation theorems state that neural networks can approximate any continuous function to any desired degree of accuracy. Nevertheless, it should be noted that this theorem does not provide any information regarding the uniqueness of the approximation. Therefore, a key limitation of the method is to ensure that the explanatory network $\mathsf{H}$ possesses sufficient constraints (both explicit and implicit) for uniqueness. The inclusion of structural symmetries (e.g. rotational invariance \cite{heider2020so}) and regularities (e.g. convexity, \cite{bengio2005convex}) will be explored in future works.

\section{Conclusion} \label{secc::conclusion}
In the present work, an enhancement of the recently introduced framework of Physically-Guided Neural Networks with Internal Variables (PGNNIV), has been proposed, not suffering from output scalability and the curse of dimensionality. The improvement has been proposed in terms of reduction of computational complexity, achieving also more noise tolerance, less overfitting and exploiting knowledge transfer to unveil new materials behavior to understand their internal structure. The method has been tested with a simple although illustrative case as it is the nonlinear diffusion equation, and is freely and openly available in a public repository.

The state of the art PGNNIV model has been modified by substituting certain parts of the predictive network (the sub-network of the complete model with the largest number of trainable parameters) with more scalable and less computationally demanding components. This change specifically concerns the decoding structure of the network. As surrogates for the decoder,  we propose a spectral approximation (e.g.  using the Fourier basis), the use of an \emph{a posteriori} orthogonal basis computed by means of the Propoer Orthogonal Decomposition (POD) and a nonlinear map based on the decoder of a pretrained autoencoder. The approaches based on a spectral reconstruction achieve a statistically significant reduction in computation time when compared to the state of the art while performing similarly in terms of predictive and explanatory capacity. On the other hand, the approaches that employ POD outperform the other approaches in predictive and explanatory accuracy, noise tolerance, and overfitting management. Moreover, they are computationally slightly more efficient than the baseline model. The autoencoder-based model exhibits the worst performance for small datasets, while in scenarios with large datasets, it performs comparably to the other models and offers a substantial reduction in training time. Ultimately, the choice of model should depend on the level of previous knowledge about the system from which the data is obtained, the available computational resources, and the acceptable margin of error. If there is prior knowledge on the physics of the system, using an adapted spectral basis can be optimal. If not, purely data-driven approaches such as POD or autoencoders can be used as alternatives, particularly for the large data samples. In particular, when working with large datasets and the goal is to discover a constitutive material law without requiring strong generalizations, the autoencoder approach is appropriate, especially under limited computational resources. If a robust model with noise filtering capacities and good generalization purposes, POD approach is preferred, with the disadvantage that it cannot be directly applied to new materials or experiments. The key point is to perform a previous evaluation of the problem to face and then select the most appropriate model based on the specific context and main application objectives.

The other way to enhance PGNNIV  proposed is to leverage previously learned knowledge to new scenarios. Transfer learning and fine-tuning techniques have been applied. The fine-tuning approach yields better results than training from scratch and achieves faster convergence. Transfer learning, on the other hand, provides a little worse predictive performance but completes the training significantly faster. Once again, the choice between these two techniques depends on the goal. If a pretrained model is available, and the objective is to rapidly uncover the material constitutive equation, transfer learning is recommended, regardless of the size of the dataset. If higher precision and better generalization are required, fine-tuning is the best option. In both cases, using either approach improves the computational efficiency compared to the state of the art, that is training from scratch.

It has been demonstrated that our approach provides both predictive accuracy and the ability to discover underlying physical laws in a computationally efficient manner and without scalability problems. The models show better noise tolerance, increased interpretability (in the sense that each spectral component may be interpreted in a more transparent way), reduced overfitting, and the potential to transfer learned knowledge to downstream tasks, leading to significant savings in computational resources. Several techniques have been introduced, and their application should be tailored to the specific problem at hand. The choice of method will depend on the characteristics of the system, the available computational capacity, and which aspects (accuracy, generalization, interpretability, training speed...) are meant to be prioritized.

\clearpage

\bibliographystyle{unsrt}  
\bibliography{references}

\end{document}